\documentclass[11pt]{article}

\usepackage[preprint]{acl}
\usepackage{times}
\usepackage{latexsym}
\usepackage{todonotes}
\usepackage[T1]{fontenc}

\usepackage[utf8]{inputenc}

\usepackage{microtype}

\usepackage{inconsolata}

\usepackage{graphicx}
\usepackage{titlesec}

\usepackage[shortlabels,inline]{enumitem}
\usepackage{tikz,lipsum}
\usepackage[most]{tcolorbox}
\usepackage{verbatim}
\usepackage{amssymb}
\usepackage{minted}
\usepackage{fancyvrb}
\usepackage{ragged2e}
\usepackage[dvipsnames]{xcolor}
\usepackage{amsmath}
\usepackage{booktabs}
\usepackage{tabularray}
\usepackage{arydshln}
\usepackage{stmaryrd}
\usepackage{marvosym}
\usepackage{makecell}
\usepackage{multirow}
\usepackage{longtable}
\usepackage{adjustbox}
\usepackage{dsfont}
\usepackage{xspace}
\usepackage{colortbl}
\usepackage{makecell}
\usepackage{multicol}
\usepackage{float}
\usepackage{placeins}
\usepackage{algorithm}
\usepackage{algpseudocode}
\usepackage{algorithmicx}
\usepackage{colortbl}
\usepackage{tablefootnote}

\usepackage{cleveref}
\crefname{section}{\S}{\S}
\crefname{table}{Table}{Tables}
\crefname{figure}{Fig.}{Figs.}
\crefname{algorithm}{Alg.}{}
\crefname{ALC@unique}{Line}{Lines}
\crefname{equation}{Eq.}{Eqs.}
\crefname{appendix}{App.}{Apps.}
\crefformat{section}{\S#2#1#3} 
\usepackage{soul}


\usepackage{todonotes}

\newcommand{\method}{\textsc{DecomposeRL}}
\newcommand{\supported}{\textsc{Supported}}
\newcommand{\refuted}{\textsc{Refuted}}
\newcommand{\nei}{\textsc{Not Enough Information}}

\NewDocumentCommand{\prompt}{O{} +m}{%
\begin{center}
\begin{tcolorbox}[
    width=0.94\columnwidth,
    coltitle=white,
    colframe=black,
    colback=black!5!white,
    boxrule=1pt,
    enhanced jigsaw,
    breakable,
    pad at break*=2mm,
    left=2pt,
    right=2pt,
    top=2pt,
    bottom=2pt,
    fontupper=\small,
    fontlower=\small,
    title={#1},
]
#2
\end{tcolorbox}
\end{center}
}

\newcommand{\promptvar}[1]{{\color{cyan}\{#1\}}}

\RequirePackage{etoolbox}
\RequirePackage{pgf}
\RequirePackage[table]{xcolor}

\definecolor{high}{HTML}{B5CCF7}
\definecolor{low}{HTML}{FFFFFF}

\newtcolorbox[auto counter,number within=section]{promptbox}[2][]{
    enhanced jigsaw,
    breakable,
    parbox=false,
    pad at break*=2mm,
    colframe=black!80,
    colback=gray!4!white,
    coltitle=white,
    colbacktitle=black!80,
    fontupper=\footnotesize,
    fontlower=\footnotesize,
    fonttitle=\small\bfseries\sffamily,
    boxrule=0.5pt,
    arc=2pt,
    left=6pt, right=6pt, top=4pt, bottom=4pt,
    title={Prompt~\thetcbcounter: #2},
    label={#1},
}

\titlespacing*{\paragraph}{0pt}{0.5ex}{0.5em}

\title{\method{}: Learning to Ask Useful, Informative, and Diverse Questions for Semi-Supervised, Traceable Claim Verification}

\newcommand{\DataReleaseURL}[0]{\url{https://dipta007.github.io/DecomposeRL}}

\author{
\textbf{Shubhashis Roy Dipta, Ankur Padia, Francis Ferraro} \\
  Department of Computer Science and Electrical Engineering\\
  University of Maryland Baltimore County\\
  Baltimore, MD 21250 USA \\
  \texttt{\{sroydip1,pankur1,ferraro\}@umbc.edu} \\
}

\begin{document}
\maketitle
\begin{abstract}
Claim verification splits between \textit{end-to-end classifiers} that are accurate but yields no inspectable traces, and \textit{decomposition-based methods} produce inspectable traces but lag performance on benchmark datasets.
We propose \method{} an accurate claim-verifier that produce inspectable traces. 
\method{} frames decomposition as an RL policy trained with GRPO and a multi-faceted reward ensemble, enabling both fully supervised and semi-supervised learning from unlabeled claims.
\method{} addresses the prohibitive training cost of GRPO with a data-curation funnel that distills $115$K fact-verification claims into a compact, learning-signal-dense subset of $5$K claims.
We show that a \method{}-$7$B policy trained with full supervision on only \textasciitilde 5K curated claims achieves $86.3$ in-domain and $69.8$ out-of-domain balanced accuracy across $11$ claim-verification benchmarks containing biomedical, political, scientific, and general-domain claims. Despite being $4{\times}$ smaller, it matches $32$B baselines and GPT-$4.1$-mini, and it further outperforms baselines in a semi-supervised setting with only $10\%$ labeled claims data.
\footnote{\DataReleaseURL}
\end{abstract}

\section{Introduction}
\label{sec:intro}
\begin{figure}[!ht]
    \centering
    \includegraphics[scale=0.8]{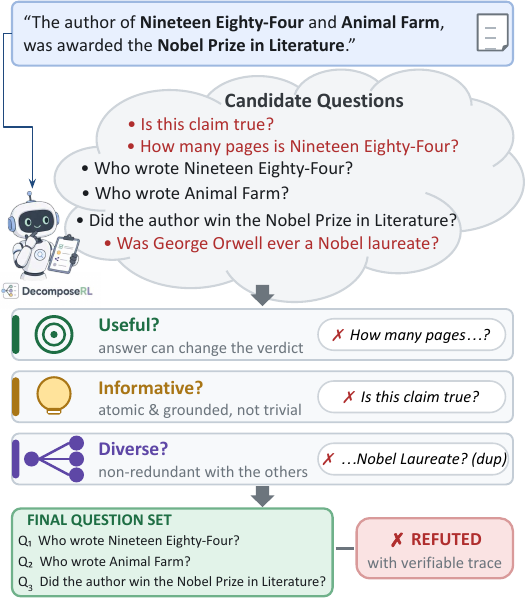}
    \caption{\textbf{What makes a question useful, informative, and diverse?} \method{} addresses this along three reward axes (full reward stack in \cref{sec:method:rewards}; full trace for this claim in \cref{fig:trace_orwell}) that filter candidate questions and output the surviving questions into an auditable trace. \emph{Useful} ($R_\text{nec}$): the answer can change the verdict; \emph{Informative} ($R_\text{joint}$): atomic, answerable, and grounded in evidence; \emph{Diverse} ($R_\text{div}$): non-redundant with the remaining questions. The three surviving questions ($Q_1$--$Q_3$) yield the correct \refuted{} verdict i.e. George Orwell was never awarded the Nobel Prize.}
\label{fig:intro_teaser}
\vspace{-4mm}
\end{figure}

Automated claim verification aims to determine the validity of a given claim and is an important task to curb the spread of misinformation. The field has split into two directions, classification and decomposition, each with complementary blind spots.

On one side, end-to-end classifiers achieve strong accuracy on claim verification at low inference cost, but remain opaque as  they emit a verdict with no inspectable traces. This opacity is problematic in high-stakes domains, e.g., biomedical literature triage~\citep{wadden2020fact}, political fact-checking~\citep{chen2022generating}, and scientific peer review~\citep{schlichtkrull2023averitec}, where the user reasonably wants to \emph{inspect} which parts of the claim were checked against the evidence, and why the verdict came out the way it did.

On the other side, decomposition-based methods \citep{press2023measuring,khot2023decomposed,chen2022generating} 
break a claim into atomic sub-questions and answer each sub-question from the evidence to compose a verdict. These methods provided inspectable traces but rely on brittle prompting to break down a claim with no training signal that measures fitness of the decomposition~\citep{press2023measuring,khot2023decomposed,zhou2023leasttomost}, require expensive high quality annotation to fine-tune an imitation-based decomposers \citep{chen2022generating}, or fail to match end-to-end classifiers performance on existing benchmarks (\cref{sec:experiment}).

In this paper, we propose \method{}, a reinforcment based multi-faceted reward ensemble approach to verify claims that produces inspectable traces while maintaining better performance. 
\method{} uses Qwen2.5-Instruct\footnote{We use an instruct-model over a reasoning variant, as it more easily supports custom structured output formats.}~\citep{yang2024qwen25} as the policy model to decompose a given claim into questions and shapes each question by its effect on the verification. The policy is trained using Group Relative Policy Optimization (GPRO) with multi-faceted reward ensemble to iteratively capture meaningful signals for the model to generate answerable and atomic question, and collectively sufficient to verify the claim either as \supported{} / \refuted{} (\cref{fig:intro_teaser}). Compared to prior approaches, \method{} keeps reinforcement training tractable with a novel \emph{data curation funnel}  to distill existing heterogeneous, noisy claim-verification corpora into a small learning-signal-dense subset. Moreover, during training, \method{} allows to combine labeled and unlabeled claims, which is critical when gold annotations are expensive and slow to collect. The proposed reward ensemble supports a semi-supervised training to score unlabeled claims with per-prompt majority-vote pseudo-labels, demonstrating that decomposition can be improved even when only a fraction of claims have a ground truth verdict.

In this paper, we make following contributions:

\begin{enumerate}
    \item A multi-faceted reward ensemble that captures multiple dimension of decomposition with per-question necessity via leave-one-out and question-set-level contribution via a joint multiplicative reward (\cref{sec:method:rewards}).
    
    \item A self-consistency reward variant to enable semi-supervised training for unlabeled claim  in settings where annotated data is scarce using intra-prompt agreement
    (\cref{sec:method:selfsup}).
    
    \item A computationally efficient data curation funnel for RL training that combines existing claim-verification corpora and distill into learning-signal-dense subset effectively using only \textasciitilde 4\% of training data
    and performing better than recent strong baselines (\cref{sec:data}, \cref{sec:experiment}).
    \item Across $11$ diverse claim-verification benchmarks, \method{}-7B outperforms comparable-size decomposition-based and end-to-end fact-checkers, matches models up to $4\times$ larger and proprietary systems, and produces traceable verification traces (\cref{sec:experiment}).
\end{enumerate}

\section{Data Curation Funnel}
\label{sec:data}

\begin{figure}[t]
\centering
\includegraphics[scale=0.8]{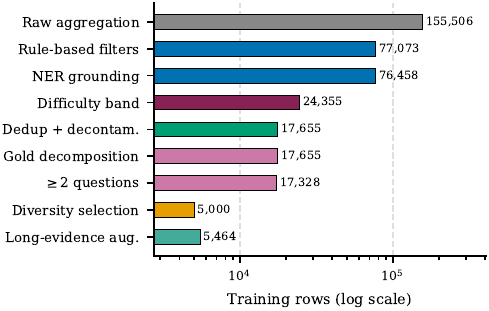}
\caption{\textbf{Data-curation funnel.}
Cumulative training-row count after each stage of the pipeline (\cref{sec:data}).}
\label{fig:data_funnel}
\vspace{-2mm}
\end{figure}

\method{} uses \emph{reward} as training signal to train a policy. 
Hence, the model's learning signal is proportional to the informativeness of the claim: a trivially-easy claim teaches the policy nothing, a mislabeled claim teaches the wrong thing, and a near-duplicate claim teaches nothing new.
We therefore apply a multi-stage data curation distillation funnel (\cref{fig:data_funnel}) that aggregates training splits from $14$ existing claim-verification corpora (\textasciitilde 155K claims) and distill them to a curated small learning-signal-dense subset of $5{,}464$ (\textasciitilde 4\%) training claims.\footnote{We will release all datasets upon acceptance.} Dataset statistics of each stage are reported in \cref{tab:source_distribution} (\cref{sec:appendix:data_sources}).
\subsection{Source Aggregation}
\label{sec:data:sources}

We pool the training splits of $14$ public claim-verification corpora into a single $(c, d, \ell^\star)$ pool: LLM-AggreFact~\citep{tang2024minicheck}, Ex-FEVER~\citep{thorne2018fever}, FEVEROUS \citep{FactExtractionVERification_2021a}, FoolMeTwice \citep{FoolMeTwice_2021}, HoVer~\citep{ho2020constructing}, FaviQA \citep{FaVIQFActVerification_2022a}, PubHealth~\citep{kotonya2020explainable}, PubHealth-Tab \citep{PubHealthTabPublicHealth_2022a}, AmbiFC \citep{AmbiFCFactCheckingAmbiguous_2024a}, WiCE \citep{WiCERealWorldEntailment_2023b}, SciFact~\citep{wadden2020fact}, SciTab \citep{SCITABChallengingBenchmark_2023b}, ClaimDecomp~\citep{chen2022generating}, and PubMedClaim \citep{PubMedQADatasetBiomedical_2019}.
Following \citet{tang2024minicheck}, we convert each corpus's native verdict scheme into either \supported{} or \refuted.
The resulting pool contains \textasciitilde 155K raw training instances that are highly heterogeneous in claim style, evidence length, and source-specific noise.

\subsection{Rule-Based Filtering}
\label{sec:data:filter:rule}

We discard claims with fewer than three evidence passages, fewer than $200$ tokens (too short to decompose meaningfully), or more than $10$k tokens (prohibitive for training compute). We further remove claims whose claim-to-evidence lexical overlap exceeds a threshold, as the evidence in such cases is essentially a paraphrase of the claim and verification reduces to trivial matching. Finally, claims with fewer than two named entities (e.g.,\ ``This is true.'') carry no learning signal and are discarded using a union of science and general-domain NER models (\cref{sec:data:filter:ner}).

\subsection{Difficulty-Based Filtering}
\label{sec:data:filter:diff}

Modern LLM-based fact-checkers~\citep{tang2024minicheck} correctly verify a substantial fraction of public-benchmark claims with high confidence. 
Such claims often do not require decomposition, while low-confidence predictions tend to expose cases where the source label itself may be noisy or incorrect~\citep{lee2022deduplicating}.
We use the MiniCheck-7B verifier~\citep{tang2024minicheck} to score each claim evidence pair, convert its probability to a label-aligned confidence score $p\in[0,1]$, and keep only claims with $0.3 \leq p \leq 0.8$ resulting in $52$k claims ($68.1\%$ of the $76$k passed by the previous stage).

\subsection{Deduplication and Decontamination}
\label{sec:data:filter:dedup}

We remove duplicate claims with MinHash-based locality-sensitive hashing~\citep{broder1997resemblance, broder2000nearduplicate} at $0.7$ Jaccard threshold, a technique similar to find near-duplicate used in large-scale LM-pretraining deduplication~\citep{lee2022deduplicating, grattafiori2024llama3, dipta2026ganitllmdifficultyawarebengalimathematical}.
Such technique catches near-duplicate claims but fails to identify paraphrased claim pairs. Hence we perform two semantic pass (a) \emph{intra-train deduplication} and (b) \emph{hold-out set decontamination}\footnote{We used \texttt{text-embedding-3-large} to obtain embeddings for each claim}.
We perform intra-train deduplication with a greedy pass that retains the first claim and removes subsequent claims whose cosine similarity to it is ${\geq}0.70$, following the threshold used in recent large-scale LM data work.~\citep{lee2022deduplicating, grattafiori2024llama3}.

Hold-out set decontamination mirrors the three-pronged decontamination protocol pioneered by~\citet{brown2020gpt3} and tightened with semantic checks in subsequent LM releases~\citep{grattafiori2024llama3}. We remove any training claim that matches a hold-out claim either by MinHash@$0.7$ or cosine similarity ${\geq}0.90$.
Lower value of MinHash threshold help to aggressively remove from training to enforce diversity and higher value of cosine similarity help to preserve hold-out distribution.

\subsection{Silver Decomposition}
\label{sec:data:filter:decomp}
We filter out claims requiring fewer than two questions, as meaningful decomposition requires at least two sub-questions. The questions are generated with \texttt{gpt-5-mini} using the claim, evidence, and rubric (\cref{sec:appendix:prompt:decompose}). Note that, we do not perform any instruction finetuning on these decompositions.

\subsection{Diversity Selection}
\label{sec:data:filter:diversity}

After deduplication and decomposition, the pool has \textasciitilde 17K claims and directly training a policy on it using GPRO is prohibitively compute-inefficient as every GRPO roll out incurs multiple LLM-as-a-judge reward calls per question. To reduce the training data size, we subsample to a budget of 5K claims under three constraints: \textbf{(i)}~per-label balance (50\%~\supported, 50\% \refuted); \textbf{(ii)}~per-source budgets allocated proportionally to $\sqrt{n_s}$, where $n_s$ is the in-pool count of claim evidence pair $s$: a square-root smoothing that softens the distribution between strict uniform and strict proportional similar to the widely used rebalancing in multilingual LM training~\citep{aharoni2019massively, conneau2020unsupervised}; and \textbf{(iii)}~within each bucket (label, claim, evidence), we maximize embedding-space diversity 
with Facility-Location objective,
\begin{equation}
    f(S)=\sum_{i\in V}\max_{j\in S}\langle c_i,c_j\rangle,
    \label{eq:facility_location}
\end{equation}
where $V$ denotes the dataset, $S$ denotes sampled set, $c$ denotes the claim. We have used lazy-greedy selection~\citep{MultilevelFacilityLocation_2015,minoux1978accelerated,mirzasoleiman2015lazier}. This objective favors representatives that cover the full embedding space of the claim, rather than its dense interior. Because it is monotone submodular, lazy-greedy gives a $(1{-}1/e)$ approximation guarantee~\citep{nemhauser1978submodular} while avoiding exhaustive search~\citep{wei2014submodular,wei2015submodularity}. The maximum-similarity form also limits outlier influence with an anomalous point may be selected once but contributes little thereafter. \cref{app:sampling_ablation} confirms that replacing submodular selection with uniform random sampling under the same budget degrades downstream accuracy, particularly on out-of-domain benchmarks; a structural comparison is provided in \cref{app:selector_comparison}.
\subsection{Long-Evidence Augmentation}
\label{sec:data:filter:aug}
The curated 5K claim set has a median evidence length of \textasciitilde{}500 tokens.
Training a policy only on shorter context causes distributions-shift at evaluation as the model would have barely seen claim with larger context.
To increase long-context coverage, we add all claims not already in the curated set whose evidence length is ${\geq}3$K tokens. This produces $464$ additional long-evidence claims, with average and maximum evidence lengths of $4.7$K and $9.8$K tokens, respectively.
\subsection{Final Distilled Training Dataset}
The augmented final pool contains $5{,}464$ claims with $51.2\%$ \supported{} / $48.8\%$ \refuted, with average silver-decomposition length of 2.63, and an evidence-length between 112 and 9,813 tokens.

\section{\method{}}
\label{sec:method}

\begin{figure*}[!htbp]
    \centering
    {\setlength{\fboxsep}{2pt}%
     \setlength{\fboxrule}{0.4pt}%
     \fcolorbox{gray!40}{white}{\includegraphics[width=\dimexpr\linewidth-2\fboxsep-2\fboxrule\relax]{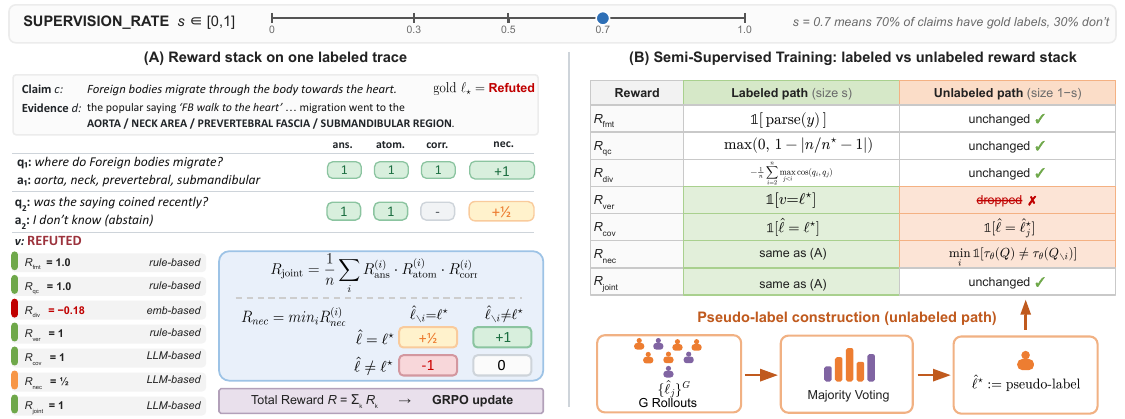}}}
    \caption{%
    \textbf{The \method{} reward ensemble and semi-supervised training.}
    Given a claim $c$ and evidence $d$, the policy $\pi_\theta$ produces a trace of $n$ question--answer cycles $(q_i, a_i)$ and a verdict $v$.
    \textbf{(A)} Seven rewards with heterogeneous evaluators: \emph{deterministic} -- format $R_\text{fmt}$, verification $R_\text{ver}$, question count $R_\text{qc}$; \emph{embedding-based} -- diversity $R_\text{div}$ (Maximal Margin Relevance over $\{q_i\}$); \emph{LLM-as-a-judge} -- coverage $R_\text{cov}$ (can the judge recover $\ell^\star$ from $\{a_i\}$ alone?), necessity $R_\text{nec}$ (leave-one-out four-state matrix: necessary / redundant / neutral / harmful, aggregated as $\min_i$), and joint quality $R_\text{joint}$ (answerability $\times$ atomicity $\times$ correctness per question, averaged over the trace). All seven sum into a single scalar for GRPO.
    \textbf{(B)} Supervision rate $s$ partitions claims into labeled (fraction $s$) and unlabeled ($1{-}s$). On the unlabeled path: $R_\text{ver}$ is dropped; $R_\text{cov}$ uses a self-consistency pseudo-label $\hat\ell^\star{=}\operatorname{mode}(\{\hat\ell_j\}_{j=1}^{G})$ from the $G$ rollouts; $R_\text{nec}$ falls back to a binary variant. The other four rewards are label-free and unchanged. ``$-$'' = factor dropped from $R_\text{joint}$ for abstentions (\cref{sec:method:joint}).
    }
    \label{fig:decomposerl_method}
    \vspace{-2mm}
\end{figure*}

\method{} frame claim verification as an iterative \emph{question--answer} (QA) decomposition problem (\S\ref{sec:method:task}) and train a language model to produce high-quality verification traces with Group Relative Policy Optimization \citep[GRPO;][]{shao2024deepseekmath}.
\method{} optimizes over a reward ensemble of seven complementary signals aiming distinct properties of a high-quality decomposition: format adherence, verdict correctness, coverage, diversity, and length constraints (\cref{sec:method:rewards}), along with two novel complementary formulations that jointly capture per-question and question-set-level quality -- a \emph{leave-one-out necessity} reward (\cref{sec:method:necessity}) and a \emph{joint multiplicative quality} reward (\cref{sec:method:joint}) to support self-consistency reward mechanism for semi-supervised learning (\cref{sec:method:selfsup}).

\subsection{Claim Verification as Iterative QA}
\label{sec:method:task}

Given a claim $c$ and an evidence document $d$, the policy $\pi_\theta$ produces a below structured trace 
\begin{equation}
\tau=\bigl(t_0,\underbrace{q_1,a_1,t_1}_{\text{cycle }1},\ldots,
\underbrace{q_n,a_n,t_n}_{\text{cycle }n},v\bigr),
\label{eq:trajectory}
\end{equation}
delimited by XML tags: an initial $\langle\textsc{think}\rangle$ block ($t_0$) that decomposes the claim into sub-claims and notes ambiguous terms; a sequence of $n{\geq}2$ question--answer cycles where each question $q_i$ targets one atomic sub-claim and $a_i$ answers it using only $d$ and a final $\langle\textsc{verification}\rangle$ block $v\in\{\supported, \refuted\}$. 
\method{} keeps the number of cycles $n$ flexible, continuing to generate until the final verification block is generated.
An iterative reward ensemble helps supervise policy decomposition into \emph{which} questions to ask and \emph{how many} to ask.

\subsection{Reward Ensemble}
\label{sec:method:rewards}

Quality of a decomposition depends on multi-dimensional and no single reward captures it: a trace can arrive at the right verdict for the wrong reasons, ask perfectly atomic questions that are unrelated to the claim, or hide a hallucinated answer behind well-formed output.
A flat sum of per-question quality scores is also insufficient, as it rewards questions that appear useful in isolation but add no evidence or support an incorrect verdict.
We therefore design the reward ensemble around two principles:
\textbf{(i) Heterogeneous Granularity:} claim-level (coverage) and sub-question-level (necessity and joint reward); \textbf{(ii) Heterogeneous Evaluators:} deterministic (format, question count, verification), embedding-based (diversity), and LLM-as-a-judge (coverage, necessity and joint reward).
\cref{fig:decomposerl_method}(A) shows a representative trace from our trained $7$B policy together with each reward value. Below are the list of rewards:

\paragraph{(1) Format} ($R_\text{fmt}$) is the fraction of well formatted structural conditions of the iterative-QA schema \cref{eq:trajectory} i.e well-formed XML, Q$\rightarrow$A alternation, and a valid final verification label, and is the basic reward to support downstream rewards.

\paragraph{(2) Verification} ($R_\text{ver}$) is direct outcome anchor that prevents the policy from optimizing proxy rewards while drifting from the end task.

\paragraph{(3) Question count} ($R_\text{qc}$) is a triangular kernel ratio $r{=}n/n^\star$, where $n$ and $n^\star$is a number decomposition from the policy and  \cref{sec:data:filter:decomp}, and is $\max(0,\, 1{-}|r{-}1|)$, peaking at $r{=}1$ and vanishing for $r{\geq}2$.
\paragraph{(4) Diversity} ($R_\text{div}$) penalizes redundancy across $\{q_1,\dots,q_n\}$ via a maximal marginal relevance score \citep{UseMMRDiversitybased_1998}, computed as  
$-\frac1n\sum\limits_{i=2}^{n}\underset{j<i}{\max}\,\cos(q_i,q_j)$
over \texttt{Qwen3-Embedding-8B} embeddings.
\paragraph{(5) Coverage} ($R_\text{cov}$) collects all answers $\{a_1,\dots,a_n\}$ and asks the LLM-as-a-judge to predict a verdict 
$\hat\ell\in\{\supported, \refuted\}$ from the answers and the claim alone, without the original document (prompt in \cref{sec:appendix:prompt:coverage}) and compared with ground truth, $R_\text{cov} {=} \mathds{1}[\hat\ell{=}\ell^\star]$.
Coverage determines if the decomposition is sufficient i.e if the ground truth label cannot be recovered from the answers alone, then the decomposition has missed something important.

\paragraph{(6) Necessity via Leave-One-Out ($R_\text{nec}$)}\label{sec:method:necessity}
To determine the necessity of a generated question, we design per-question necessity criteria with leaving-one-out strategy where a question is necessary if removing it changes the verdict. Asking the LLM judge ``how relevant is the question $q_i$ to claim $c$?'' for each of the questions wrongly scored \textasciitilde 0.9 for almost every question including those that are tangentially related to the claim, and those that produce the incorrect verdict.

For each $c_i$ we run on the full answer set  and on the leave-one-out set $A_{\setminus i}{=}\{a_j: j{\neq}i\}$, to obtain verdicts $\hat\ell$ and $\hat\ell_{\setminus i}$ to compare with the ground truth $\ell^\star$, and score each question on a 2$\times$2 matrix:
\[
R_\text{nec}^{(i)} = \begin{cases}
+1 & \hat\ell{=}\ell^\star,\;\hat\ell_{\setminus i}{\neq}\ell^\star \quad\text{(necessary)}\\
+\tfrac{1}{2} & \hat\ell{=}\ell^\star,\;\hat\ell_{\setminus i}{=}\ell^\star \quad\text{(redundant)}\\
0 & \hat\ell{\neq}\ell^\star,\;\hat\ell_{\setminus i}{\neq}\ell^\star \quad\text{(neutral)}\\
-1 & \hat\ell{\neq}\ell^\star,\;\hat\ell_{\setminus i}{=}\ell^\star \quad\text{(harmful)}.
\end{cases}
\]
As shown in \cref{fig:decomposerl_method}(A), on the running trace, $q_{1}$ receives \texttt{+1} because dropping it breaks the judge's reconstruction of the ground truth \refuted{} verdict, while the abstention $q_2$ receives  $\mathtt{+\tfrac{1}{2}}$ because removing it leaves the verdict unchanged.
The \emph{harmful} case (\texttt{-1}) is the most informative, to identify if a question has misleads the verifier. Also, since it assigns negative rewards, it can push the policy to \emph{remove} questions rather than refine them.
We aggregate individual question score to a trace-level scalar as $R_{\mathrm{nec}}=\min\limits_i R_{\mathrm{nec}}^{(i)}$.
Using average aggregation would smooths the worst-case question against the rest of the trace and lets a single harmful question hide behind several necessary ones. However, the minimum aggregation forfeits any reward for a trace containing even one harmful question.

\subsubsection*{(7) Joint Multiplicative Quality}
\label{sec:method:joint}

Quality of a generated question is determined with three criteria: if it is answerable from the evidence (\emph{answerability}), if it isolates a single sub-claim (\emph{atomicity}), and if its answer is faithful to the document (\emph{correctness}).
We score each criterion with a LLM-as-a-judge-based sub-signal and combine the three into a multiplicative trace-level reward.

\paragraph{(7a) Answerability ($R_\text{ans}^{(i)}$).}
Prompts the judge with $(d, q_i)$ and asks whether $q_i$ is fully answerable from $d$ alone (see \cref{sec:appendix:prompt:question_checker}).
The score is $1$ if the judge declares $q_i$ answerable from $d$, and $0$ otherwise.

\paragraph{(7b) Atomicity ($R_\text{atom}^{(i)}$).}
Instead of using  atomicity prompts for $0$--$10$, we found checklist based atomicity more robust. Checklist (\cref{sec:appendix:prompt:atomicity}) has five binary criterion: the question must be (i) an actual question, (ii) single-focus, (iii) free of compound conjunctions joining sub-claims, (iv) verifiable (yes/no or specific factual answer), and (v) grounded in claim-specific entities and score fraction passed and average across questions as signal.

\paragraph{(7c) Answer correctness ($R_\text{corr}^{(i)}$).}
Takes each question--answer pair $(q_i, a_i)$ and asks the judge if the answer contradicts, conatins extrinsic information to  $d$ (\cref{sec:appendix:prompt:answer_checker}).

\paragraph{Multiplicative composite.}
Summing the above score lets the policy compensate for any one failed criterion by inflating the others. Hence, we combine them \emph{multiplicatively} per question and average over the trace:
\begin{equation}
R_\text{joint} = \frac{1}{n}\sum_{i=1}^{n} R_\text{ans}^{(i)} \cdot R_\text{atom}^{(i)} \cdot R_\text{corr}^{(i)}.
\label{eq:joint_multiplicative_reward}
\end{equation}
A single failure on any criterion drives the per-question term to zero.
For answer abstentions (``I don't know''), we drop the undefined $R_\text{corr}^{(i)}$ factor and score the question as $R_\text{ans}^{(i)} \cdot R_\text{atom}^{(i)}$, so an honest abstention earns reward proportional to the question's quality rather than being penalized for the missing answer.
On the running trace in \cref{fig:decomposerl_method}(A), each cycle scores the per-question maximum, $q_1$ via the full three-factor product and $q_2$ via the abstention rule, giving $R_\text{joint}{=}1$ for the trace.

\paragraph{Composition and amortization.}
The seven rewards combine additively into a single trace-level scalar $R(\tau){=}\sum_k R_k(\tau)$, which GRPO normalizes against group rollout before the policy update.

\subsection{Semi-Supervision With Reward Ensemble}
\label{sec:method:selfsup}

Claim-verification labels are expensive as real-world claims require expert annotations, making it difficult to scale \method{} to new domains without training data. 
To support training a policy without labeled data, \method{} uses the reward ensemble to extract the gradient from claims.
As shown in the \cref{fig:decomposerl_method}(B) \method{} support supervision rate, $s\in[0,1]$ to partition the training pool into a \emph{labeled path} of expected fraction $s$ and an \emph{unlabeled path} of expected fraction $1-s$.
The split is computed once per claim before training begins, ensuring that each claim is assigned to the same partition across epochs and runs.
\begin{table*}[!htbp]
\centering
\small
\setlength{\tabcolsep}{4pt}
\renewcommand{\arraystretch}{1.05}
\resizebox{\textwidth}{!}{%
\begin{tabular}{@{}l@{\hspace{6pt}}cccccccccc@{\hspace{8pt}}ccc@{}}
\toprule
 & \multicolumn{10}{c}{\textbf{In-Domain (9 datasets)}} & \multicolumn{3}{c}{\textbf{Out-of-Domain (2 datasets)}} \\
\cmidrule(lr){2-11} \cmidrule(lr){12-14}
\textbf{Method} & \textbf{FEVER} & \textbf{ClaimDecomp} & \textbf{HoVer} & \textbf{FEVEROUS} & \textbf{WiCE} & \textbf{Ex-FEVER} & \textbf{PubHealth} & \textbf{PubMedClaim} & \textbf{FoolMeTwice} & \textbf{Avg} & \textbf{CoverBench} & \textbf{LLM-AggreFact} & \textbf{Avg} \\
\midrule
\rowcolor{gray!15}
\multicolumn{14}{@{}l}{\textit{Base policy (Qwen2.5-7B-Instruct, prompted)}} \\
Simple & \underline{72.7} & 94.9 & 71.0 & \textbf{93.5} & 83.2 & 82.7 & 84.2 & \underline{84.1} & 86.6 & 83.7 & 52.5 & 74.9 & 63.7 \\
CoT & 70.0 & 95.5 & 70.9 & 92.2 & 85.6 & \underline{83.8} & 83.8 & 83.2 & 85.0 & 83.3 & 59.7 & 77.2 & 68.5 \\
\addlinespace[2pt]
\rowcolor{gray!15}
\multicolumn{14}{@{}l}{\textit{Fine-tuned classifier}} \\
MiniCheck-7B & 69.9 & 77.5 & \underline{73.8} & 89.2 & \underline{87.2} & 82.9 & 76.3 & 83.0 & 84.5 & 80.5 & 54.6 & \textbf{80.3} & 67.5 \\
\addlinespace[2pt]
\rowcolor{gray!15}
\multicolumn{14}{@{}l}{\textit{Decomposition-style methods (Qwen2.5-7B-Instruct backbone)}} \\
Self-Ask & 66.5 & 92.7 & 66.9 & 91.9 & 82.5 & 71.7 & 84.2 & 82.6 & 82.8 & 80.2 & 56.9 & 77.1 & 67.0 \\
Decomposed Prompting & 65.5 & 95.3 & 69.0 & 91.9 & 85.0 & 78.0 & 85.7 & 82.5 & 84.1 & 81.9 & 55.3 & 76.2 & 65.8 \\
HiSS & 67.7 & 92.8 & 70.2 & 92.7 & 83.6 & 82.4 & 79.2 & 77.0 & 84.5 & 81.1 & 58.3 & 75.7 & 67.0 \\
FOLK & 65.0 & 90.8 & 68.2 & 91.0 & 83.6 & 80.2 & 80.5 & 77.8 & 83.1 & 80.0 & 53.8 & 75.6 & 64.7 \\
ProgramFC & 60.5 & 92.9 & 65.9 & 88.2 & 85.4 & 74.6 & 77.4 & 74.3 & 76.9 & 77.3 & 53.1 & 73.5 & 63.3 \\
Chen-2024 & 65.4 & 91.1 & 65.3 & 87.9 & 79.6 & 73.3 & 83.3 & 79.2 & 82.3 & 78.6 & 56.8 & 70.2 & 63.5 \\
ClaimDecomp & 65.2 & 78.9 & 63.5 & 85.5 & 79.2 & 71.6 & 76.0 & 77.6 & 79.4 & 75.2 & 52.1 & 71.6 & 61.9 \\
QACheck & 65.4 & 97.3 & 59.1 & 92.7 & 83.0 & 65.4 & \textbf{91.0} & 78.0 & 81.6 & 79.3 & 52.8 & 68.9 & 60.9 \\
\midrule
\textbf{\method{} ($s{=}1.0$)} & \textbf{74.1} & \textbf{98.6} & \textbf{76.4} & \underline{93.1} & 86.5 & \textbf{87.6} & \underline{87.5} & \textbf{85.5} & \textbf{87.7} & \textbf{86.3} & \textbf{62.5} & 77.0 & \textbf{69.8} \\
\textbf{\method{} ($s{=}0.1$)} & 71.4 & \underline{98.1} & 70.4 & 92.9 & \textbf{87.9} & 82.6 & 86.9 & 83.9 & \underline{87.1} & \underline{84.6} & \underline{60.6} & \underline{78.7} & \underline{69.7} \\
\bottomrule
\end{tabular}}
\caption{\textbf{Balanced accuracy (\%) at the $7$B parameter scale on 9 in-domain and 2 out-of-domain datasets.} 
All decomposition-based methods use \texttt{Qwen2.5-7B-Instruct}; MiniCheck-$7$B \citep{tang2024minicheck} is a separately distilled $7$B fact-checker. \method{} ($s{=}1.0$) is the full-supervision policy, and \method{} ($s{=}0.1$) is the semi-supervised variant trained with only $10\%$ ground truth labels (\cref{sec:exp:semisup}). \textbf{Bold} denotes the best result and \underline{underline} the runner-up. \method{}~($s{=}1.0$) achieves the best overall average on both aggregates.
\method{}~($s{=}0.1$) beats every baseline on both aggregates and is the runner-up on six other datasets.
}
\label{tab:main_combined_7b}
\vspace{-2mm}
\end{table*}

Due to the presence of unlabeled claims, out of seven rewards, Verification $R_\text{ver}$, Coverage $R_\text{cov}$ and Necessity $R_{nec}$ become void as it cannot be computed without ground truth label $\ell^\star$ while the remaining four are independent of $\ell^\star$ and operate identically on labeled and unlabeled claims. To overcome, we compute Coverage and Necessity as follows:

\paragraph{Coverage via intra-prompt agreement.}
To recover a coverage signal on the unlabeled path we replace the ground truth label with a \emph{self-consistency pseudo-label}~\citep{wang2023selfconsistency} built from the policy's own rollouts.
GRPO already samples $G$ trajectories per prompt, all sharing the same claim and document.
We re-use those $G$ verdicts $\{\hat\ell_j\}_{j=1}^{G}$, take the majority vote 
$\hat\ell^\star=\operatorname*{mode}\limits_{j}\hat\ell_j$
as the pseudo-label, and re-score each rollout against the group: $R_\text{cov}^{(j)}{=}\mathds{1}[\hat\ell_j{=}\hat\ell^\star]$.
The signal rewards \emph{intra-prompt agreement} rather than absolute correctness, but the two are tightly correlated when the claim is verifiable from the document~\citep{wang2023selfconsistency} as most rollouts converge on the same verdict and the pseudo-label tracks the ground truth label.

\paragraph{Necessity becomes relative.}
Necessity (\cref{sec:method:necessity}) matrix is replaced with the binary variant where 
$R_\text{nec}^{(i)}{=}1$ if removing $a_i$ changes the 
verdict, else $0$.
Such rewards setting still measures contribution rather than relevance, but is resolved against the policy's reconstruction of the verdict instead of a ground truth label.

\section{Experiments} \label{sec:experiment}

\paragraph{Model and training.}
\label{sec:exp:setup}
For the policy model, we use \texttt{Qwen2.5-7B-Instruct}~\citep{yang2024qwen25}, and train with GRPO~\citep{shao2024deepseekmath} using LoRA~\citep{hu2022lora} on the $5{,}464$ curated claims as described in \cref{sec:data}.
We used \texttt{Qwen3-32B} as reward judge served via vLLM~\citep{kwon2023vllm} under deterministic decoding, with cached judge responses so the same prompt receives the same reward across epochs and resumes.
Additional implementation details are reported in \cref{sec:appendix:impl}.

\paragraph{Evaluation benchmarks.}
We evaluate on $11$ held-out claim-verification real-world benchmarks datasets with $9$ \emph{in-domain} and $2$ \emph{out-of-domain}. Section \cref{sec:appendix:benchmarks} contains the full benchmark list and per-corpus domain grouping.

\begin{table*}[!t]
\centering
\small
\setlength{\tabcolsep}{4pt}
\renewcommand{\arraystretch}{1.05}
\resizebox{\textwidth}{!}{%
\begin{tabular}{@{}l@{\hspace{6pt}}cccccccccc@{\hspace{8pt}}ccc@{}}
\toprule
 & \multicolumn{10}{c}{\textbf{In-Domain (9 datasets)}} & \multicolumn{3}{c}{\textbf{Out-of-Domain}} \\
\cmidrule(lr){2-11} \cmidrule(lr){12-14}
\textbf{Method} & \textbf{FEVER} & \textbf{ClaimDecomp} & \textbf{HoVer} & \textbf{FEVEROUS} & \textbf{WiCE} & \textbf{Ex-FEVER} & \textbf{PubHealth} & \textbf{PubMedClaim} & \textbf{FoolMeTwice} & \textbf{Avg} & \textbf{CoverBench} & \textbf{LLM-AggreFact} & \textbf{Avg} \\
\midrule
Simple @ 3B & 71.5 & 94.0 & 63.7 & 89.0 & 73.7 & 82.0 & 81.2 & 78.7 & 79.3 & 79.2 & 51.3 & 74.0 & 62.7 \\
Simple @ 7B & \underline{72.7} & 94.9 & 71.0 & \underline{93.5} & 83.2 & 82.7 & 84.2 & 84.1 & 86.6 & 83.7 & 52.5 & 74.9 & 63.7 \\
Decomposed Prompting @ 14B & 71.1 & \textbf{100.0} & 75.0 & 90.9 & \underline{89.0} & 83.4 & 86.7 & 85.3 & 88.3 & 85.5 & 61.3 & \underline{79.3} & 70.3 \\
Decomposed Prompting @ 32B & 68.6 & \underline{100.0} & 76.2 & 93.2 & \textbf{91.3} & 85.1 & \underline{86.8} & \textbf{87.4} & \underline{90.3} & \underline{86.5} & \underline{64.2} & \textbf{79.4} & \underline{71.8} \\
Self-Ask @ GPT-4.1-mini & 70.9 & 100.0 & \textbf{76.7} & \textbf{93.5} & 87.2 & \textbf{88.3} & 86.4 & \underline{87.1} & \textbf{91.1} & \textbf{86.8} & \textbf{68.6} & 78.9 & \textbf{73.8} \\
\midrule
\textbf{\method{} @ 7B (Ours)} & \textbf{74.1} & 98.6 & \underline{76.4} & 93.1 & 86.5 & \underline{87.6} & \textbf{87.5} & 85.5 & 87.7 & 86.3 & 62.5 & 77.0 & 69.8 \\
\bottomrule
\end{tabular}}
\caption{\textbf{Balanced accuracy (\%) across model sizes.} For each size, we report the best-performing baseline.
\textbf{Bold} marks the best result and \underline{underline} the second best. \method{}-$7$B is within $0.5$ points of both the $32$B baseline and the frontier model on in-domain average accuracy ($86.3$ vs.\ $86.8$), and within $0.5$ points of Decomposed Prompting-$14$B on out-of-domain balanced accuracy. Overall, \method{} performance matches 4$\times$ larger and production models with an order-of-magnitude smaller policy.
}
\label{tab:main_combined_best}
\vspace{-2mm}
\end{table*}

\paragraph{Metric and checkpoint selection.}
Following \citet{tang2024minicheck}, we report \emph{balanced accuracy}, which is robust to per-benchmark label skew.
For aggregate comparison, we report the average balanced accuracy, computed as the uniform mean of the nine per-dataset scores for the in-domain setting and the mean of the two scores for the out-of-domain setting, respectively.

\paragraph{Baselines.}
We compare \method{} against three families of baselines.
\paragraph{(i) Base policy:} \texttt{Qwen2.5-7B-Instruct}, the starting model before RL, prompted directly with the claim and evidence in two variants, \emph{Simple} (verdict-only) and \emph{CoT}~\citep{wei2022chain}.
\paragraph{(ii) Fine-tuned classifier:} MiniCheck-$7$B~\citep{tang2024minicheck}, a distilled fact-checker that emits a single verdict without an inspection surface.
\paragraph{(iii) Decomposition-style methods:} Self-Ask~\citep{press2023measuring}, Decomposed Prompting~\citep{khot2023decomposed}, ClaimDecomp~\citep{chen2022generating}, HiSS~\citep{LLMbasedFactVerification_2023b}, FOLK~\citep{ExplainableClaimVerification_2023b}, ProgramFC~\citep{FactCheckingComplexClaims_2023b}, QACheck~\citep{QACheckDemonstrationSystem_2023}, and Chen-$2024$~\citep{ComplexClaimVerification_2024a} as an supervised-fine-tuned decomposer pipeline (\cref{sec:related}).
For scale, we include \texttt{Qwen2.5-3/7/14/32B-Instruct} and \texttt{gpt-4.1-mini} as base model decompositions. 

\subsection{Main Results} \label{sec:exp:main}

\paragraph{\method{} performs better than strong baselines.}
As shown in \autoref{tab:main_combined_7b}, on average, \method{} obtains balanced accuracy of 86.3 in-domain and 69.8 out-of-domain.
Compared to decomposition style methods, \method{} gains $+4.4$ points for in-domain and $+2.8$ points for out-of-domain  over the strongest method, decomposed prompting and HiSS, respectively.
Comparing against the base policy and fine-tuned fact checker (Simple, CoT, and MiniCheck), \method{} gains $+2.6$ in-domain and $+1.3$ out-of-domain over the strongest baselines; MiniCheck trails despite being a dedicated fine-tuned fact-checker.
\method{} performs better, both in and out-of-domain, due to the multi-faceted reward ensemble that improved verdict accuracy without trading off out-of-distribution robustness.

\paragraph{\method{} performance matches frontier and $4{\times}$-larger methods.}
\autoref{tab:main_combined_best} compares the performance of  \method{} ($7$B parameter model) with the best-performing baseline at each model size ($3$B, $7$B, $14$B, $32$B) along with a frontier model (GPT 4.1 mini).
For in-domain, \method{}-$7$B ($86.3$) ties Decomposed Prompting~@~$32$B ($86.5$) and is less than a point of the GPT-$4.1$-mini Self-Ask frontier ($86.8$), despite using a policy that is $4{\times}$ smaller than the $32$B baseline and an order of magnitude smaller than the frontier model.
Considering out-of-domain, \method{}-$7$B ($69.8$) is less than a point of Decomposed Prompting~@~$14$B ($70.3$) at half the parameters, and sightly lower compared to the GPT-$4.1$-mini frontier ($73.8$).
The reward ensemble thus closes the bulk of the scale gap with parameter-efficient supervision rather than parameter count.

\paragraph{Verdicts come with an inspectable trace.}
Every \method{} verdict comes with a structured trace which includes  evidence-based questions, respective answers, and a calibrated abstention slot, which a downstream reader can audit.
\cref{app:trace_examples} shows four representative traces: clean refutation, calibrated abstention, and counting-style failure.

\subsection{Semi-Supervised Training} \label{sec:exp:semisup}

\method{} supports semi-supervised training (\cref{sec:method:selfsup}) in low label settings. We vary the supervision rate $s\in\{0.1, 0.3, 0.5, 0.7, 1.0\}$, which controls the fraction of training claims with ground-truth labels, and measure downstream accuracy on the same benchmark datasets.
The full/low ($s{=}1.0/0.1$) supervision results are in \cref{tab:main_combined_7b}; results of all $s$ are in \cref{app:supervision}.

\paragraph{$\mathbf{10\%}$ of ground truth label beats every $7$B baseline.}
At $s{=}0.1$, \method{}-$7$B reaches average balanced accuracy of $84.6$ in-domain and $69.7$ out-of-domain, thereby exceeding the strongest $7$B baselines (see \cref{tab:main_combined_7b}). 
\method{}-$7$B ($s{=}0.1$) accuracy drops  by $1.7$ and $0.1$ points for in and out-of-domain. The reward ensembles therefore carries the bulk of the learning signal.

\paragraph{Implication for label-scarce domains.}
As shown in  \cref{tab:main_combined_7b} (\method{} at $s{=}0.1$, bottom row) shows that \method{} is useful for domains where verdict annotation is expensive and scarce. Reward ensembles and pseudo-label strategy as described in \cref{sec:method:selfsup} makes it possible to maintain balanced accuracy while using only handful data. This observation is similar to previous work ~\citep{ReinforcementLearningSelfDistillation_2026, ModelCanHelp_2026a}, where the policy score its own rollouts to replace of external verdicts as a viable substitute for ground truth supervision.

\subsection{Reward Ensemble Ablation} \label{sec:exp:reward}
\begin{table}[t]
\centering
\small
\setlength{\tabcolsep}{5pt}
\renewcommand{\arraystretch}{1.05}
\begin{tabular}{@{}l cc cc@{}}
\toprule
 & \multicolumn{2}{c}{\textbf{In-Domain}} & \multicolumn{2}{c}{\textbf{Out-of-Domain}} \\
\cmidrule(lr){2-3} \cmidrule(lr){4-5}
\textbf{Variant} & \textbf{Avg} & \textbf{$\Delta$} & \textbf{Avg} & \textbf{$\Delta$} \\
\midrule
\textbf{\method{}} & \textbf{86.3} & -- & \textbf{69.8} & -- \\
\addlinespace[2pt]
\;\;$-$ Necessity       & 86.2 & $-0.1$ & 65.0 & $-4.8$ \\
\;\;$-$ Coverage        & 86.2 & $-0.1$ & 65.9 & $-3.9$ \\
\;\;$-$ Diversity       & 85.7 & $-0.6$ & 66.6 & $-3.2$ \\
\;\;$-$ Joint Quality   & 85.6 & $-0.7$ & 67.8 & $-2.0$ \\
\;\;$-$ Question Count  & 86.2 & $-0.1$ & 68.1 & $-1.7$ \\
\bottomrule
\end{tabular}
\caption{\textbf{Reward ensemble ablation.} Each row removes one reward. In-domain is robust ($\le$ 0.7) while out-of-domain drops up by 4.8 (Necessity). Full breakdown in \cref{app:full_reward_ablation}.}
\label{tab:reward_ablation}
\vspace{-4mm}
\end{table}

The reward ensemble contains seven signals (\cref{sec:method:rewards}). Format ($R_\text{fmt}$) and Verification ($R_\text{ver}$) are structural: removing either collapses training because the policy loses its output-shape constraint. We ablate the remaining five by removing one at a time while keeping everything else fixed (\cref{tab:reward_ablation}).
For in-domain Joint Multiplicative Quality and Diversity are important as removing them cost the balanced accuracy to drop by 0.6 and 0.7 points while three of the five ablations cost $\le 0.1$.
For out-of-domain, every reward removal degrades the out-of-domain balanced accuracy.
The leave-one-out necessity reward (\cref{sec:method:necessity}) is the single most impactful signal for generalization, removing it causes the model balanced accuracy to drop by 4.8 points.
Coverage and diversity are the next most impactful signals to drop by 3.9 and 3.2 balanced accuracy, consistent with their shared role in ensuring the decomposition collectively covers the claim. No individual reward is dominated by the others; the full ensemble is necessary for out-of-domain robustness.

\section{Conclusion}

We present \method{}, a reinforcement-learning approach that trains the claim decomposer to produce atomic questions that are simultaneously useful for verification and auditable by a human. The multi-faceted reward ensemble which include format adherence, length constraint, diversity, outcome, coverage, necessity, and joint reward provide enough signal to train a $7$B policy that matches systems $4{\times}$ its size and an order-of-magnitude larger frontier model across $11$ benchmarks, using only ${\sim}5$k curated training claims. Finally, \method{} extends to semi-supervised training, using only $10\%$ gold-label supervision while maintaining both in-domain and out-of-domain accuracy above every same-scale baseline.

\section*{Limitations} \label{sec:limitations}

\paragraph{Reliance on pre-retrieved evidence.}
\method{} assumes a fixed evidence document per claim and does not retrieve new evidence during the trace; reported numbers therefore measure decomposition quality given evidence, not end-to-end fact-checking quality.
This isolates the contribution of the reward ensemble from retrieval quality, and the trace structure is retriever-agnostic -- any retriever can be dropped in as a front end.

\paragraph{Dependence on the LLM judge.}
Five of seven rewards are scored by a Qwen$3$-$32$B judge, so judge blind spots can in principle be inherited under sustained reward pressure~\citep{gao2023scaling, zheng2023judging}; the judge is also the dominant training cost.
Heterogeneous evaluators (two rule-based, one embedding-based) prevent collapse onto a single judge, the multiplicative composite (\cref{sec:method:joint}) and leave-one-out necessity (\cref{sec:method:necessity}) require multiple judge calls to agree -- and the judge is never called at inference time.

\paragraph{Binary verdict head.}
\method{} generates a $2$-way verdict (\supported{} / \refuted), matching prior decomposition baselines; ``not enough information'' is expressible at the per-question level via calibrated abstention but not at the trace level.
The coverage judge already produces a $3$-way verdict internally, so extending the head is a one-line change rather than a methodological obstacle.

\section*{Acknowledgment}
Some experiments were conducted on the UMBC HPCF, supported by the National Science Foundation under Grant No. CNS-1920079. %
This material is based on research supported by DARPA for the SciFy program under agreement number HR00112520301. %
The U.S. Government is authorized to reproduce and distribute reprints for Governmental purposes notwithstanding any copyright notation thereon. The views and conclusions contained herein are those of the authors and should not be interpreted as necessarily representing the official policies or endorsements, either express or implied, of DARPA or the U.S. Government.

\bibliography{dipta_rebib}

@article{AmbiFCFactCheckingAmbiguous_2024a,
    author = {Glockner, Max  and
Stali{\=u}nait{\.e}, Ieva  and
Thorne, James  and
Vallejo, Gisela  and
Vlachos, Andreas  and
Gurevych, Iryna},
    journal = {Transactions of the Association for Computational Linguistics},
    title = {{A}mbi{FC}: Fact-Checking Ambiguous Claims with Evidence},
    year = {2024}
}

@inproceedings{aharoni2019massively,
  title={Massively multilingual neural machine translation},
  author={Aharoni, Roee and Johnson, Melvin and Firat, Orhan},
  booktitle={Proceedings of the 2019 Conference of the North American Chapter of the Association for Computational Linguistics: Human Language Technologies, Volume 1 (Long and Short Papers)},
  pages={3874--3884},
  year={2019}
}

@inproceedings{broder1997resemblance,
    author = {Broder, Andrei Z.},
    booktitle = {Proceedings of the Compression and Complexity of Sequences (SEQUENCES)},
    title = {On the resemblance and containment of documents},
    year = {1997}
}

@inproceedings{broder2000nearduplicate,
    author = {Broder, Andrei Z.},
    journal = {Combinatorial Pattern Matching (CPM)},
    title = {Identifying and filtering near-duplicate documents},
    year = {2000}
}

@inproceedings{brown2020gpt3,
    author = {Tom B. Brown and
Benjamin Mann and
Nick Ryder and
Melanie Subbiah and
Jared Kaplan and
Prafulla Dhariwal and
Arvind Neelakantan and
Pranav Shyam and
Girish Sastry and
Amanda Askell and
Sandhini Agarwal and
Ariel Herbert{-}Voss and
Gretchen Krueger and
Tom Henighan and
Rewon Child and
Aditya Ramesh and
Daniel M. Ziegler and
Jeffrey Wu and
Clemens Winter and
Christopher Hesse and
Mark Chen and
Eric Sigler and
Mateusz Litwin and
Scott Gray and
Benjamin Chess and
Jack Clark and
Christopher Berner and
Sam McCandlish and
Alec Radford and
Ilya Sutskever and
Dario Amodei},
    booktitle = {Advances in Neural Information Processing Systems 33: Annual Conference
on Neural Information Processing Systems 2020, NeurIPS 2020, December
6-12, 2020, virtual},
    editor = {Hugo Larochelle and
Marc'Aurelio Ranzato and
Raia Hadsell and
Maria{-}Florina Balcan and
Hsuan{-}Tien Lin},
    title = {Language Models are Few-Shot Learners},
    year = {2020}
}

@inproceedings{chen2022generating,
    author = {Chen, Jifan  and
Sriram, Aniruddh  and
Choi, Eunsol  and
Durrett, Greg},
    booktitle = {Proceedings of the 2022 Conference on Empirical Methods in Natural Language Processing},
    editor = {Goldberg, Yoav  and
Kozareva, Zornitsa  and
Zhang, Yue},
    title = {Generating Literal and Implied Subquestions to Fact-check Complex Claims},
    year = {2022}
}

@inproceedings{ComplexClaimVerification_2024a,
    author = {Chen, Jifan  and
Kim, Grace  and
Sriram, Aniruddh  and
Durrett, Greg  and
Choi, Eunsol},
    booktitle = {Proceedings of the 2024 Conference of the North American Chapter of the Association for Computational Linguistics: Human Language Technologies (Volume 1: Long Papers)},
    editor = {Duh, Kevin  and
Gomez, Helena  and
Bethard, Steven},
    title = {Complex Claim Verification with Evidence Retrieved in the Wild},
    year = {2024}
}

@inproceedings{conneau2020unsupervised,
    author = {Conneau, Alexis  and
Khandelwal, Kartikay  and
Goyal, Naman  and
Chaudhary, Vishrav  and
Wenzek, Guillaume  and
Guzm{\'a}n, Francisco  and
Grave, Edouard  and
Ott, Myle  and
Zettlemoyer, Luke  and
Stoyanov, Veselin},
    booktitle = {Proceedings of the 58th Annual Meeting of the Association for Computational Linguistics},
    editor = {Jurafsky, Dan  and
Chai, Joyce  and
Schluter, Natalie  and
Tetreault, Joel},
    title = {Unsupervised Cross-lingual Representation Learning at Scale},
    year = {2020}
}

@misc{deepseek2025r1,
    author = {D Guo and D Yang and H Zhang and J Song and P Wang and others},
    title = {Deepseek-r1: Incentivizing reasoning capability in llms via reinforcement learning},
    year = {2025}
}

@misc{dipta2026ganitllmdifficultyawarebengalimathematical,
    author = {Shubhashis Roy Dipta and Khairul Mahbub and Nadia Najjar},
    journal = {ArXiv preprint},
    title = {GanitLLM: Difficulty-Aware Bengali Mathematical Reasoning through Curriculum-GRPO},
    year = {2026}
}

@misc{dipta2026pa3policyawareagentalignment,
    author = {Shubhashis Roy Dipta and Daniel Bis and Kun Zhou and Lichao Wang and Benjamin Z. Yao and Chenlei Guo and Ruhi Sarikaya},
    journal = {ArXiv preprint},
    title = {PA3: Policy-Aware Agent Alignment through Chain-of-Thought},
    year = {2026}
}

@inproceedings{dua2022successive,
    author = {Dua, Dheeru  and
Gupta, Shivanshu  and
Singh, Sameer  and
Gardner, Matt},
    booktitle = {Proceedings of the 2022 Conference on Empirical Methods in Natural Language Processing},
    editor = {Goldberg, Yoav  and
Kozareva, Zornitsa  and
Zhang, Yue},
    title = {Successive Prompting for Decomposing Complex Questions},
    year = {2022}
}

@inproceedings{EXFEVERDatasetMultihop_2024a,
    author = {Ma, Huanhuan and Xu, Weizhi and Wei, Yifan and Chen, Liuji and Wang, Liang and Liu, Qiang and Wu, Shu and Wang, Liang},
    booktitle = {Findings of the {Association} for {Computational} {Linguistics}: {ACL} 2024},
    editor = {Ku, Lun-Wei and Martins, Andre and Srikumar, Vivek},
    shorttitle = {{EX}-{FEVER}},
    title = {{EX}-{FEVER}: {A} {Dataset} for {Multi}-hop {Explainable} {Fact} {Verification}},
    urldate = {2026-02-14},
    year = {2024}
}

@inproceedings{ExplainableClaimVerification_2023b,
    author = {Wang, Haoran  and
Shu, Kai},
    booktitle = {Findings of the Association for Computational Linguistics: EMNLP 2023},
    editor = {Bouamor, Houda  and
Pino, Juan  and
Bali, Kalika},
    title = {Explainable Claim Verification via Knowledge-Grounded Reasoning with Large Language Models},
    year = {2023}
}

@inproceedings{FactCheckingComplexClaims_2023b,
    author = {Pan, Liangming  and
Wu, Xiaobao  and
Lu, Xinyuan  and
Luu, Anh Tuan  and
Wang, William Yang  and
Kan, Min-Yen  and
Nakov, Preslav},
    booktitle = {Proceedings of the 61st Annual Meeting of the Association for Computational Linguistics (Volume 1: Long Papers)},
    editor = {Rogers, Anna  and
Boyd-Graber, Jordan  and
Okazaki, Naoaki},
    title = {Fact-Checking Complex Claims with Program-Guided Reasoning},
    year = {2023}
}

@inproceedings{FactExtractionVERification_2021a,
    author = {Aly, Rami  and
Guo, Zhijiang  and
Schlichtkrull, Michael Sejr  and
Thorne, James  and
Vlachos, Andreas  and
Christodoulopoulos, Christos  and
Cocarascu, Oana  and
Mittal, Arpit},
    booktitle = {Proceedings of the Fourth Workshop on Fact Extraction and VERification (FEVER)},
    editor = {Aly, Rami  and
Christodoulopoulos, Christos  and
Cocarascu, Oana  and
Guo, Zhijiang  and
Mittal, Arpit  and
Schlichtkrull, Michael  and
Thorne, James  and
Vlachos, Andreas},
    title = {The Fact Extraction and {VER}ification Over Unstructured and Structured information ({FEVEROUS}) Shared Task},
    year = {2021}
}

@inproceedings{FaVIQFActVerification_2022a,
    author = {Park, Jungsoo  and
Min, Sewon  and
Kang, Jaewoo  and
Zettlemoyer, Luke  and
Hajishirzi, Hannaneh},
    booktitle = {Proceedings of the 60th Annual Meeting of the Association for Computational Linguistics (Volume 1: Long Papers)},
    editor = {Muresan, Smaranda  and
Nakov, Preslav  and
Villavicencio, Aline},
    title = {{F}a{VIQ}: {FA}ct Verification from Information-seeking Questions},
    year = {2022}
}

@inproceedings{FoolMeTwice_2021,
    author = {Eisenschlos, Julian  and
Dhingra, Bhuwan  and
Bulian, Jannis  and
B{\"o}rschinger, Benjamin  and
Boyd-Graber, Jordan},
    booktitle = {Proceedings of the 2021 Conference of the North American Chapter of the Association for Computational Linguistics: Human Language Technologies},
    editor = {Toutanova, Kristina  and
Rumshisky, Anna  and
Zettlemoyer, Luke  and
Hakkani-Tur, Dilek  and
Beltagy, Iz  and
Bethard, Steven  and
Cotterell, Ryan  and
Chakraborty, Tanmoy  and
Zhou, Yichao},
    title = {Fool Me Twice: Entailment from {W}ikipedia Gamification},
    year = {2021}
}

@inproceedings{gao2023scaling,
    author = {Leo Gao and
John Schulman and
Jacob Hilton},
    booktitle = {International Conference on Machine Learning, {ICML} 2023, 23-29 July
2023, Honolulu, Hawaii, {USA}},
    editor = {Andreas Krause and
Emma Brunskill and
Kyunghyun Cho and
Barbara Engelhardt and
Sivan Sabato and
Jonathan Scarlett},
    series = {Proceedings of Machine Learning Research},
    title = {Scaling Laws for Reward Model Overoptimization},
    year = {2023}
}

@article{grattafiori2024llama3,
    author = {Grattafiori, Aaron and Dubey, Abhimanyu and Jauhri, Abhinav and Pandey, Abhinav and Kadian, Abhishek and Al-Dahle, Ahmad and Letman, Aiesha and Mathur, Akhil and Schelten, Alan and Vaughan, Alex and others},
    journal = {ArXiv preprint},
    title = {The {L}lama 3 Herd of Models},
    year = {2024}
}

@inproceedings{ho2020constructing,
    author = {Ho, Xanh  and
Duong Nguyen, Anh-Khoa  and
Sugawara, Saku  and
Aizawa, Akiko},
    booktitle = {Proceedings of the 28th International Conference on Computational Linguistics},
    editor = {Scott, Donia  and
Bel, Nuria  and
Zong, Chengqing},
    title = {Constructing A Multi-hop {QA} Dataset for Comprehensive Evaluation of Reasoning Steps},
    year = {2020}
}

@article{honnibal2020spacy,
    author = {Honnibal, Matthew and Montani, Ines and Van Landeghem, Sofie and Boyd, Adriane},
    title = {spaCy: Industrial-strength Natural Language Processing in Python},
    year = {2020}
}

@inproceedings{hu2022lora,
    author = {Edward J. Hu and
Yelong Shen and
Phillip Wallis and
Zeyuan Allen{-}Zhu and
Yuanzhi Li and
Shean Wang and
Lu Wang and
Weizhu Chen},
    booktitle = {The Tenth International Conference on Learning Representations, {ICLR}
2022, Virtual Event, April 25-29, 2022},
    title = {LoRA: Low-Rank Adaptation of Large Language Models},
    year = {2022}
}

@inproceedings{khot2023decomposed,
    author = {Tushar Khot and
Harsh Trivedi and
Matthew Finlayson and
Yao Fu and
Kyle Richardson and
Peter Clark and
Ashish Sabharwal},
    booktitle = {The Eleventh International Conference on Learning Representations,
{ICLR} 2023, Kigali, Rwanda, May 1-5, 2023},
    title = {Decomposed Prompting: {A} Modular Approach for Solving Complex Tasks},
    year = {2023}
}

@inproceedings{kotonya2020explainable,
    author = {Kotonya, Neema  and
Toni, Francesca},
    booktitle = {Proceedings of the 2020 Conference on Empirical Methods in Natural Language Processing (EMNLP)},
    editor = {Webber, Bonnie  and
Cohn, Trevor  and
He, Yulan  and
Liu, Yang},
    title = {Explainable Automated Fact-Checking for Public Health Claims},
    year = {2020}
}

@inproceedings{kwon2023vllm,
    author = {Kwon, Woosuk and Li, Zhuohan and Zhuang, Siyuan and Sheng, Ying and Zheng, Lianmin and Yu, Cody Hao and Gonzalez, Joseph E. and Zhang, Hao and Stoica, Ion},
    booktitle = {Proceedings of the 29th Symposium on Operating Systems Principles},
    title = {Efficient Memory Management for Large Language Model Serving with {P}aged{A}ttention},
    year = {2023}
}

@article{lanham2023measuring,
    author = {Lanham, Tamera and Chen, Anna and Radhakrishnan, Ansh and Steiner, Benoit and Denison, Carson and Hernandez, Danny and Li, Dustin and Durmus, Esin and Hubinger, Evan and Kernion, Jackson and others},
    journal = {ArXiv preprint},
    title = {Measuring faithfulness in chain-of-thought reasoning},
    year = {2023}
}

@inproceedings{lee2022deduplicating,
    author = {Lee, Katherine  and
Ippolito, Daphne  and
Nystrom, Andrew  and
Zhang, Chiyuan  and
Eck, Douglas  and
Callison-Burch, Chris  and
Carlini, Nicholas},
    booktitle = {Proceedings of the 60th Annual Meeting of the Association for Computational Linguistics (Volume 1: Long Papers)},
    editor = {Muresan, Smaranda  and
Nakov, Preslav  and
Villavicencio, Aline},
    title = {Deduplicating Training Data Makes Language Models Better},
    year = {2022}
}

@inproceedings{lightman2024lets,
    author = {Hunter Lightman and
Vineet Kosaraju and
Yuri Burda and
Harrison Edwards and
Bowen Baker and
Teddy Lee and
Jan Leike and
John Schulman and
Ilya Sutskever and
Karl Cobbe},
    booktitle = {The Twelfth International Conference on Learning Representations,
{ICLR} 2024, Vienna, Austria, May 7-11, 2024},
    title = {Let's Verify Step by Step},
    year = {2024}
}

@inproceedings{liu2023geval,
    author = {Liu, Yang  and
Iter, Dan  and
Xu, Yichong  and
Wang, Shuohang  and
Xu, Ruochen  and
Zhu, Chenguang},
    booktitle = {Proceedings of the 2023 Conference on Empirical Methods in Natural Language Processing},
    editor = {Bouamor, Houda  and
Pino, Juan  and
Bali, Kalika},
    title = {{G}-Eval: {NLG} Evaluation using Gpt-4 with Better Human Alignment},
    year = {2023}
}

@inproceedings{LLMbasedFactVerification_2023b,
    author = {Zhang, Xuan  and
Gao, Wei},
    booktitle = {Proceedings of the 13th International Joint Conference on Natural Language Processing and the 3rd Conference of the Asia-Pacific Chapter of the Association for Computational Linguistics (Volume 1: Long Papers)},
    editor = {Park, Jong C.  and
Arase, Yuki  and
Hu, Baotian  and
Lu, Wei  and
Wijaya, Derry  and
Purwarianti, Ayu  and
Krisnadhi, Adila Alfa},
    title = {Towards {LLM}-based Fact Verification on News Claims with a Hierarchical Step-by-Step Prompting Method},
    year = {2023}
}

@inproceedings{min2023factscore,
    author = {Min, Sewon  and
Krishna, Kalpesh  and
Lyu, Xinxi  and
Lewis, Mike  and
Yih, Wen-tau  and
Koh, Pang  and
Iyyer, Mohit  and
Zettlemoyer, Luke  and
Hajishirzi, Hannaneh},
    booktitle = {Proceedings of the 2023 Conference on Empirical Methods in Natural Language Processing},
    editor = {Bouamor, Houda  and
Pino, Juan  and
Bali, Kalika},
    title = {{FA}ct{S}core: Fine-grained Atomic Evaluation of Factual Precision in Long Form Text Generation},
    year = {2023}
}

@incollection{minoux1978accelerated,
    author = {Minoux, Michel},
    booktitle = {Optimization Techniques},
    title = {Accelerated greedy algorithms for maximizing submodular set functions},
    year = {1978}
}

@inproceedings{mirzasoleiman2015lazier,
    author = {Baharan Mirzasoleiman and
Ashwinkumar Badanidiyuru and
Amin Karbasi and
Jan Vondr{\'{a}}k and
Andreas Krause},
    booktitle = {Proceedings of the Twenty-Ninth {AAAI} Conference on Artificial Intelligence,
January 25-30, 2015, Austin, Texas, {USA}},
    editor = {Blai Bonet and
Sven Koenig},
    title = {Lazier Than Lazy Greedy},
    year = {2015}
}

@misc{jacovi2024coverbench,
    author = {A Jacovi and M Ambar and E Ben and others},
    title = {Coverbench: A challenging benchmark for complex claim verification},
    year = {2024}
}

@article{MultilevelFacilityLocation_2015,
    author = {Ortiz Astorquiza, Camilo and Contreras, Ivan and Laporte, Gilbert},
    journal = {European Journal of Operational Research},
    title = {Multi-level facility location as the maximization of a submodular set function},
    year = {2015}
}

@article{nemhauser1978submodular,
    author = {Nemhauser, George L. and Wolsey, Laurence A. and Fisher, Marshall L.},
    journal = {Mathematical Programming},
    number = {1},
    title = {An analysis of approximations for maximizing submodular set functions---{I}},
    year = {1978}
}

@inproceedings{neumann2019scispacy,
    author = {Neumann, Mark  and
King, Daniel  and
Beltagy, Iz  and
Ammar, Waleed},
    booktitle = {Proceedings of the 18th BioNLP Workshop and Shared Task},
    editor = {Demner-Fushman, Dina  and
Cohen, Kevin Bretonnel  and
Ananiadou, Sophia  and
Tsujii, Junichi},
    title = {{S}cispa{C}y: Fast and Robust Models for Biomedical Natural Language Processing},
    year = {2019}
}

@inproceedings{ouyang2022training,
    author = {Long Ouyang and
Jeffrey Wu and
Xu Jiang and
Diogo Almeida and
Carroll L. Wainwright and
Pamela Mishkin and
Chong Zhang and
Sandhini Agarwal and
Katarina Slama and
Alex Ray and
John Schulman and
Jacob Hilton and
Fraser Kelton and
Luke Miller and
Maddie Simens and
Amanda Askell and
Peter Welinder and
Paul F. Christiano and
Jan Leike and
Ryan Lowe},
    booktitle = {Advances in Neural Information Processing Systems 35: Annual Conference
on Neural Information Processing Systems 2022, NeurIPS 2022, New Orleans,
LA, USA, November 28 - December 9, 2022},
    editor = {Sanmi Koyejo and
S. Mohamed and
A. Agarwal and
Danielle Belgrave and
K. Cho and
A. Oh},
    title = {Training language models to follow instructions with human feedback},
    year = {2022}
}

@inproceedings{press2023measuring,
    author = {Press, Ofir  and
Zhang, Muru  and
Min, Sewon  and
Schmidt, Ludwig  and
Smith, Noah  and
Lewis, Mike},
    booktitle = {Findings of the Association for Computational Linguistics: EMNLP 2023},
    editor = {Bouamor, Houda  and
Pino, Juan  and
Bali, Kalika},
    title = {Measuring and Narrowing the Compositionality Gap in Language Models},
    year = {2023}
}

@inproceedings{PubMedQADatasetBiomedical_2019,
    author = {Jin, Qiao  and
Dhingra, Bhuwan  and
Liu, Zhengping  and
Cohen, William  and
Lu, Xinghua},
    booktitle = {Proceedings of the 2019 Conference on Empirical Methods in Natural Language Processing and the 9th International Joint Conference on Natural Language Processing (EMNLP-IJCNLP)},
    editor = {Inui, Kentaro  and
Jiang, Jing  and
Ng, Vincent  and
Wan, Xiaojun},
    title = {{P}ub{M}ed{QA}: A Dataset for Biomedical Research Question Answering},
    year = {2019}
}

@inproceedings{QACheckDemonstrationSystem_2023,
    author = {Pan, Liangming  and
Lu, Xinyuan  and
Kan, Min-Yen  and
Nakov, Preslav},
    booktitle = {Proceedings of the 2023 Conference on Empirical Methods in Natural Language Processing: System Demonstrations},
    editor = {Feng, Yansong  and
Lefever, Els},
    title = {{QAC}heck: A Demonstration System for Question-Guided Multi-Hop Fact-Checking},
    year = {2023}
}

@article{radhakrishnan2023question,
    author = {Radhakrishnan, Ansh and Nguyen, Karina and Chen, Anna and Chen, Carol and Denison, Carson and Hernandez, Danny and Durmus, Esin and Hubinger, Evan and Kernion, Jackson and Lukosuite, Kamile and others},
    journal = {ArXiv preprint},
    title = {Question Decomposition Improves the Faithfulness of Model-Generated Reasoning},
    year = {2023}
}

@inproceedings{rafailov2023direct,
    author = {Rafael Rafailov and
Archit Sharma and
Eric Mitchell and
Christopher D. Manning and
Stefano Ermon and
Chelsea Finn},
    booktitle = {Advances in Neural Information Processing Systems 36: Annual Conference
on Neural Information Processing Systems 2023, NeurIPS 2023, New Orleans,
LA, USA, December 10 - 16, 2023},
    editor = {Alice Oh and
Tristan Naumann and
Amir Globerson and
Kate Saenko and
Moritz Hardt and
Sergey Levine},
    title = {Direct Preference Optimization: Your Language Model is Secretly a
Reward Model},
    year = {2023}
}

@inproceedings{schlichtkrull2023averitec,
    author = {Michael Schlichtkrull and
Zhijiang Guo and
Andreas Vlachos},
    booktitle = {Advances in Neural Information Processing Systems 36: Annual Conference
on Neural Information Processing Systems 2023, NeurIPS 2023, New Orleans,
LA, USA, December 10 - 16, 2023},
    editor = {Alice Oh and
Tristan Naumann and
Amir Globerson and
Kate Saenko and
Moritz Hardt and
Sergey Levine},
    title = {AVeriTeC: {A} Dataset for Real-world Claim Verification with Evidence
from the Web},
    year = {2023}
}

@article{schulman2017proximal,
    author = {Schulman, John and Wolski, Filip and Dhariwal, Prafulla and Radford, Alec and Klimov, Oleg},
    journal = {ArXiv preprint},
    title = {Proximal policy optimization algorithms},
    year = {2017}
}

@inproceedings{SCITABChallengingBenchmark_2023b,
    author = {Lu, Xinyuan  and
Pan, Liangming  and
Liu, Qian  and
Nakov, Preslav  and
Kan, Min-Yen},
    booktitle = {Proceedings of the 2023 Conference on Empirical Methods in Natural Language Processing},
    editor = {Bouamor, Houda  and
Pino, Juan  and
Bali, Kalika},
    title = {{SCITAB}: A Challenging Benchmark for Compositional Reasoning and Claim Verification on Scientific Tables},
    year = {2023}
}

@inproceedings{setlur2025rewarding,
  title={Rewarding progress: Scaling automated process verifiers for llm reasoning},
  author={Setlur, Amrith and Nagpal, Chirag and Fisch, Adam and Geng, Xinyang and Eisenstein, Jacob and Agarwal, Rishabh and Agarwal, Alekh and Berant, Jonathan and Kumar, Aviral},
  booktitle={International Conference on Learning Representations},
  volume={2025},
  pages={60808--60838},
  year={2025}
}

@article{shao2024deepseekmath,
    author = {Shao, Zhihong and Wang, Peiyi and Zhu, Qihao and Xu, Runxin and Song, Junxiao and Bi, Xiao and Zhang, Haowei and Zhang, Mingchuan and Li, Y.K. and Wu, Y. and Guo, Daya},
    journal = {ArXiv preprint},
    title = {{DeepSeekMath}: Pushing the Limits of Mathematical Reasoning in Open Language Models},
    year = {2024}
}

@inproceedings{tang2024minicheck,
    author = {Tang, Liyan and Laban, Philippe and Durrett, Greg},
    booktitle = {Proceedings of the 2024 Conference on Empirical Methods in Natural Language Processing},
    title = {{MiniCheck}: Efficient Fact-Checking of {LLM}s on Grounding Documents},
    year = {2024}
}

@inproceedings{thorne2018fever,
    author = {Thorne, James  and
Vlachos, Andreas  and
Christodoulopoulos, Christos  and
Mittal, Arpit},
    booktitle = {Proceedings of the 2018 Conference of the North {A}merican Chapter of the Association for Computational Linguistics: Human Language Technologies, Volume 1 (Long Papers)},
    editor = {Walker, Marilyn  and
Ji, Heng  and
Stent, Amanda},
    title = {{FEVER}: a Large-scale Dataset for Fact Extraction and {VER}ification},
    year = {2018}
}

@article{trivedi2022musique,
    author = {Trivedi, Harsh  and
Balasubramanian, Niranjan  and
Khot, Tushar  and
Sabharwal, Ashish},
    editor = {Roark, Brian  and
Nenkova, Ani},
    journal = {Transactions of the Association for Computational Linguistics},
    title = {{M}u{S}i{Q}ue: Multihop Questions via Single-hop Question Composition},
    year = {2022}
}

@inproceedings{UseMMRDiversitybased_1998,
    author = {Carbonell, Jaime and Goldstein, Jade},
    booktitle = {Proceedings of the 21st annual international {ACM} {SIGIR} conference on {Research} and development in information retrieval},
    isbn = {978-1-58113-015-7},
    series = {{SIGIR} '98},
    title = {The use of {MMR}, diversity-based reranking for reordering documents and producing summaries},
    urldate = {2026-05-21},
    year = {1998}
}

@inproceedings{wadden2020fact,
    author = {Wadden, David  and
Lin, Shanchuan  and
Lo, Kyle  and
Wang, Lucy Lu  and
van Zuylen, Madeleine  and
Cohan, Arman  and
Hajishirzi, Hannaneh},
    booktitle = {Proceedings of the 2020 Conference on Empirical Methods in Natural Language Processing (EMNLP)},
    editor = {Webber, Bonnie  and
Cohn, Trevor  and
He, Yulan  and
Liu, Yang},
    title = {Fact or Fiction: Verifying Scientific Claims},
    year = {2020}
}

@inproceedings{wang2023selfconsistency,
    author = {Xuezhi Wang and
Jason Wei and
Dale Schuurmans and
Quoc V. Le and
Ed H. Chi and
Sharan Narang and
Aakanksha Chowdhery and
Denny Zhou},
    booktitle = {The Eleventh International Conference on Learning Representations,
{ICLR} 2023, Kigali, Rwanda, May 1-5, 2023},
    title = {Self-Consistency Improves Chain of Thought Reasoning in Language Models},
    year = {2023}
}

@inproceedings{wei2014submodular,
    author = {Wei, Kai and Liu, Yuzong and Kirchhoff, Katrin and Bilmes, Jeff},
    booktitle = {2014 IEEE International Conference on Acoustics, Speech and Signal Processing (ICASSP)},
    title = {Submodular subset selection for large-scale speech training data},
    year = {2014}
}

@inproceedings{wei2015submodularity,
    author = {Kai Wei and
Rishabh K. Iyer and
Jeff A. Bilmes},
    booktitle = {Proceedings of the 32nd International Conference on Machine Learning,
{ICML} 2015, Lille, France, 6-11 July 2015},
    editor = {Francis R. Bach and
David M. Blei},
    series = {{JMLR} Workshop and Conference Proceedings},
    title = {Submodularity in Data Subset Selection and Active Learning},
    year = {2015}
}

@inproceedings{wei2022chain,
    author = {Jason Wei and
Xuezhi Wang and
Dale Schuurmans and
Maarten Bosma and
Brian Ichter and
Fei Xia and
Ed H. Chi and
Quoc V. Le and
Denny Zhou},
    booktitle = {Advances in Neural Information Processing Systems 35: Annual Conference
on Neural Information Processing Systems 2022, NeurIPS 2022, New Orleans,
LA, USA, November 28 - December 9, 2022},
    editor = {Sanmi Koyejo and
S. Mohamed and
A. Agarwal and
Danielle Belgrave and
K. Cho and
A. Oh},
    title = {Chain-of-Thought Prompting Elicits Reasoning in Large Language Models},
    year = {2022}
}

@inproceedings{wei2024longform,
    author = {Jerry Wei and
Chengrun Yang and
Xinying Song and
Yifeng Lu and
Nathan Hu and
Jie Huang and
Dustin Tran and
Daiyi Peng and
Ruibo Liu and
Da Huang and
Cosmo Du and
Quoc V. Le},
    booktitle = {Advances in Neural Information Processing Systems 38: Annual Conference
on Neural Information Processing Systems 2024, NeurIPS 2024, Vancouver,
BC, Canada, December 10 - 15, 2024},
    editor = {Amir Globersons and
Lester Mackey and
Danielle Belgrave and
Angela Fan and
Ulrich Paquet and
Jakub M. Tomczak and
Cheng Zhang},
    title = {Long-form factuality in large language models},
    year = {2024}
}

@inproceedings{WiCERealWorldEntailment_2023b,
    author = {Kamoi, Ryo  and
Goyal, Tanya  and
Diego Rodriguez, Juan  and
Durrett, Greg},
    booktitle = {Proceedings of the 2023 Conference on Empirical Methods in Natural Language Processing},
    editor = {Bouamor, Houda  and
Pino, Juan  and
Bali, Kalika},
    title = {{W}i{CE}: Real-World Entailment for Claims in {W}ikipedia},
    year = {2023}
}

@article{wolfson2020break,
    author = {Wolfson, Tomer  and
Geva, Mor  and
Gupta, Ankit  and
Gardner, Matt  and
Goldberg, Yoav  and
Deutch, Daniel  and
Berant, Jonathan},
    editor = {Johnson, Mark  and
Roark, Brian  and
Nenkova, Ani},
    journal = {Transactions of the Association for Computational Linguistics},
    title = {Break It Down: A Question Understanding Benchmark},
    year = {2020}
}

@inproceedings{yang2018hotpotqa,
    author = {Yang, Zhilin  and
Qi, Peng  and
Zhang, Saizheng  and
Bengio, Yoshua  and
Cohen, William  and
Salakhutdinov, Ruslan  and
Manning, Christopher D.},
    booktitle = {Proceedings of the 2018 Conference on Empirical Methods in Natural Language Processing},
    editor = {Riloff, Ellen  and
Chiang, David  and
Hockenmaier, Julia  and
Tsujii, Jun{'}ichi},
    title = {{H}otpot{QA}: A Dataset for Diverse, Explainable Multi-hop Question Answering},
    year = {2018}
}

@article{yang2024qwen25,
    author = {Yang, An and Yang, Baosong and Zhang, Beichen and Hui, Binyuan and Zheng, Bo and Yu, Bowen and Li, Chengyuan and Liu, Dayiheng and Huang, Fei and Wei, Haoran and others},
    journal = {ArXiv preprint},
    title = {Qwen2.5 Technical Report},
    year = {2024}
}

@inproceedings{zheng2023judging,
    author = {Lianmin Zheng and
Wei{-}Lin Chiang and
Ying Sheng and
Siyuan Zhuang and
Zhanghao Wu and
Yonghao Zhuang and
Zi Lin and
Zhuohan Li and
Dacheng Li and
Eric P. Xing and
Hao Zhang and
Joseph E. Gonzalez and
Ion Stoica},
    booktitle = {Advances in Neural Information Processing Systems 36: Annual Conference
on Neural Information Processing Systems 2023, NeurIPS 2023, New Orleans,
LA, USA, December 10 - 16, 2023},
    editor = {Alice Oh and
Tristan Naumann and
Amir Globerson and
Kate Saenko and
Moritz Hardt and
Sergey Levine},
    title = {Judging LLM-as-a-Judge with MT-Bench and Chatbot Arena},
    year = {2023}
}

@inproceedings{zhou2023leasttomost,
    author = {Denny Zhou and
Nathanael Sch{\"{a}}rli and
Le Hou and
Jason Wei and
Nathan Scales and
Xuezhi Wang and
Dale Schuurmans and
Claire Cui and
Olivier Bousquet and
Quoc V. Le and
Ed H. Chi},
    booktitle = {The Eleventh International Conference on Learning Representations,
{ICLR} 2023, Kigali, Rwanda, May 1-5, 2023},
    title = {Least-to-Most Prompting Enables Complex Reasoning in Large Language
Models},
    year = {2023}
}

@inproceedings{PubHealthTabPublicHealth_2022a,
    title = {{PubHealthTab}: {A} {Public} {Health} {Table}-based {Dataset} for {Evidence}-based {Fact} {Checking}},
    booktitle = {Findings of the {Association} for {Computational} {Linguistics}: {NAACL} 2022},
    publisher = {Association for Computational Linguistics},
    author = {Akhtar, Mubashara and Cocarascu, Oana and Simperl, Elena},
}

@misc{ModelCanHelp_2026a,
    title = {A {Model} {Can} {Help} {Itself}: {Reward}-{Free} {Self}-{Training} for {LLM} {Reasoning}},
    shorttitle = {A {Model} {Can} {Help} {Itself}},
    url = {http://arxiv.org/abs/2510.18814},
    doi = {10.48550/arXiv.2510.18814},
    language = {en},
    urldate = {2026-05-19},
    publisher = {arXiv},
    author = {Li, Mengqi and Zhao, Lei and So, Anthony Man-Cho and Sun, Ruoyu and Li, Xiao},
    month = may,
    year = {2026},
    note = {arXiv:2510.18814 [cs.LG]},
}

@misc{ReinforcementLearningSelfDistillation_2026,
    title = {Reinforcement {Learning} via {Self}-{Distillation}},
    url = {http://arxiv.org/abs/2601.20802},
    doi = {10.48550/arXiv.2601.20802},
    urldate = {2026-05-15},
    publisher = {arXiv},
    author = {Hübotter, Jonas and Lübeck, Frederike and Behric, Lejs and Baumann, Anton and Bagatella, Marco and Marta, Daniel and Hakimi, Ido and Shenfeld, Idan and Buening, Thomas Kleine and Guestrin, Carlos and Krause, Andreas},
    month = feb,
    year = {2026},
    note = {arXiv:2601.20802 [cs.LG]},
}

\newpage
\appendix
\section*{Appendix}

\section{Related Work}
\label{sec:related}

\paragraph{Decomposed claim verification: prompted, supervised, untrained.}
The intuition that complex claims are best verified by breaking them into atomic sub-questions runs through both claim-verification and broader multi-hop reasoning literature.
\citet{chen2022generating} introduced ClaimDecomp, a corpus pairing claims with gold yes/no decompositions and showing that decomposed verdicts beat monolithic ones; QDMR~\citep{wolfson2020break} provides a more general structured representation for question decomposition, and FActScore~\citep{min2023factscore} formalizes atomic checking for long-form factuality.
Closer to our task, \citet{ComplexClaimVerification_2024a} (henceforth \textbf{Chen-$2024$}) embed a learned claim-decomposer inside an end-to-end fact-checking pipeline with retrieval and claim-focused summarization, training the decomposer on existing gold decompositions.
A parallel prompting line: chain-of-thought~\citep{wei2022chain}, Self-Ask~\citep{press2023measuring}, Decomposed Prompting~\citep{khot2023decomposed}, least-to-most~\citep{zhou2023leasttomost}, successive prompting~\citep{dua2022successive}, shows decompositions zero-shot, and connects to multi-hop QA datasets including HotpotQA~\citep{yang2018hotpotqa}, HoVer~\citep{ho2020constructing}, and MuSiQue~\citep{trivedi2022musique}.
Specifically for claim verification, HiSS~\citep{LLMbasedFactVerification_2023b}, FOLK~\citep{ExplainableClaimVerification_2023b}, ProgramFC~\citep{FactCheckingComplexClaims_2023b}, and QACheck~\citep{QACheckDemonstrationSystem_2023} all run iterative question-guided decomposition with in-context LLMs, differing primarily in the intermediate representation: hierarchical sub-claims, first-order-logic clauses, executable programs, and free-form question chains, respectively. \citet{radhakrishnan2023question} and \citet{lanham2023measuring} arguing that decomposition yields more faithful chains of thought than plain CoT.
These methods share the structural limitation that motivates our paper: the decomposer is either \emph{untrained} (prompted) or trained only to \emph{imitate} a gold reference (ClaimDecomp/QDMR-style), with no signal that measures whether the decomposition is actually \emph{useful} for the downstream verifier.
We retain the iterative-QA decomposition formulation but replace both supervisory regimes with rewards defined by verdict consequences.

\paragraph{End-to-end fact-checking and the traceability gap.}
Fine-tuned classifiers on FEVER~\citep{thorne2018fever}, SciFact~\citep{wadden2020fact}, and FEVEROUS dominated early benchmarks; AVeriTeC~\citep{schlichtkrull2023averitec} extended evaluation to real-world claims with retrieved evidence.
More recently, distillation-based fact-checkers like MiniCheck~\citep{tang2024minicheck} match GPT-4 accuracy at $7$B parameters on the LLM-AggreFact benchmark by training on synthetic factual-error data.
These systems are fast and accurate on standard short-claim tests, but they emit a single \supported{}/\refuted{} verdict with no inspection surface, and recent benchmarks targeting complex multi-hop and long-evidence claims (CoverBench~\citep{jacovi2024coverbench}) document a substantial residual gap to frontier models.
Our work is positioned in the space these methods leave open: we keep their accuracy target but recover an inspectable, structured trace.

\paragraph{RL for LLM reasoning: outcome vs.\ process rewards.}
RLHF~\citep{ouyang2022training} established RL with outcome rewards as a canonical alignment recipe; PPO~\citep{schulman2017proximal} and the critic-free DPO~\citep{rafailov2023direct} are the standard optimizers, and GRPO~\citep{shao2024deepseekmath} further removes the critic and made consumer-scale reasoning RL tractable, culminating in DeepSeek-R1's demonstration that pure outcome rewards can drive complex reasoning at scale~\citep{deepseek2025r1}.
Outcome-only rewards, however, give no credit assignment over intermediate steps, a known limitation~\citep{lanham2023measuring} that has motivated a parallel line of work on \emph{process reward models}: PRM800K-style human-annotated step labels~\citep{lightman2024lets}, automatic process-reward construction from rollout outcomes~\citep{setlur2025rewarding}, and self-consistency-derived rewards~\citep{wang2023selfconsistency}.
Reward models more broadly are also known to be vulnerable to over-optimization at scale~\citep{gao2023scaling}, sharpening the case for diverse, complementary reward signals.
Our reward stack contributes a different design point: rather than scoring each step's \emph{intrinsic} correctness, we score each step's \emph{causal contribution} to the final verdict, via leave-one-out necessity and a multiplicative composite that requires simultaneous step-level success on multiple axes.

\paragraph{LLM-as-judge in the loop.}
We rely on a strong LLM judge~\citep{yang2024qwen25} to score nuanced reward components (atomicity, answerability, coverage), following a body of work that validates LLMs as cheap evaluators of text quality~\citep{zheng2023judging, liu2023geval, dipta2026pa3policyawareagentalignment} and as factuality decomposers for long-form generation~\citep{min2023factscore, wei2024longform}.
But LLM judges are known to exhibit position bias, verbosity bias, and self-enhancement bias~\citep{zheng2023judging}, and reward models built on them can be exploited via reward hacking under heavy optimization pressure~\citep{gao2023scaling}.

\paragraph{Recent Advances in RL for LLM}. Recent advances in RL for LLM reasoning, particularly group-relative methods such as GRPO~\citep{shao2024deepseekmath, deepseek2025r1, dipta2026ganitllmdifficultyawarebengalimathematical}, make this technically feasible: the optimizer is off-the-shelf, and the hard problem moves from collecting more annotations to designing the right reward landscape.
\emph{Outcome-only} scoring (verdict correctness) rewards ``lucky-guess'' decompositions that arrive at the right verdict via off-topic questions and provide no credit or penalty assignment over the question-answer cycle that produced them~\citep{lightman2024lets, setlur2025rewarding}.
\emph{Per-question LLM-judge} rewards (atomicity, saliency) score each sub-question in isolation and are subject to known to LLM biases, e.g., position bias, self-consistency bias~\citep{zheng2023judging, liu2023geval} which make their signals unreliable.

\section{NER Grounding}
\label{sec:data:filter:ner}
Claims without named entities (e.g.,\ ``This is true.'') carry no learning signal for these criteria, and therefore claims with fewer than two named entities are discarded combining science and general domain. This union ensures coverage of both scientific entity mentions missed by the general model and common entities overlooked by the science model\footnote{We used (\texttt{en\_core\_sci\_lg}~\citep{neumann2019scispacy}) and (\texttt{en\_core\_web\_trf} \citep{honnibal2020spacy} for science and general domain.}. Although this stage removes only $0.8\%$ of the claims, it serves as a necessary precondition for the downstream judge-based filters.

\section{Per-Source Training Counts}
\label{sec:appendix:data_sources}

\cref{tab:source_distribution} reports per-source row counts after each stage of the data-curation pipeline described in \cref{sec:data}.
The aggregated raw pool spans $14$ corpora across Wikipedia, news, biomedical, and tabular domains; the rule-based pass, difficulty band, semantic deduplication, and submodular diversity selection each contribute their own attrition pattern, summarised here for per-source attribution rather than reproduced in the body.

\begin{table}[t]
\centering
\tiny
\setlength{\tabcolsep}{4pt}
\renewcommand{\arraystretch}{1.05}
\begin{tabular}{@{}llrrrr@{}}
\toprule
\textbf{Source} & \textbf{Domain} & \textbf{Raw} & \textbf{Diff.} & \textbf{Dedup} & \textbf{Final} \\
\midrule
Ex-FEVER          & Wiki         & 36{,}841 &  8{,}394 & 3{,}716 &    823 \\
LLM-AggreFact     & News         & 30{,}420 &  5{,}498 & 4{,}275 & 1{,}069 \\
FEVEROUS          & Wiki         & 26{,}928 &     818 &    787 &    237 \\
HoVer             & Wiki         & 18{,}171 &  2{,}533 & 2{,}106 &    559 \\
FoolMeTwice       & Wiki         & 11{,}588 &  2{,}992 & 2{,}684 &    663 \\
FaviQ-A           & Wiki         & 10{,}924 &     444 &    358 &    238 \\
PubHealth         & Health       &  9{,}096 &  2{,}418 & 2{,}357 & 1{,}044 \\
AmbiFC            & Wiki         &  4{,}912 &     303 &    176 &    122 \\
PubHealth-Tab     & Health (tab) &  2{,}664 &     224 &    191 &    139 \\
SciFact           & Biomedical   &  1{,}295 &     113 &    106 &     85 \\
SciTab            & Sci. (tab)   &     868  &     226 &    184 &    161 \\
WiCE              & Wiki         &     785  &     233 &    229 &    182 \\
ClaimDecomp       & Politics     &     569  &      40 &     40 &     40 \\
PubMedClaim       & Biomedical   &     445  &     119 &    119 &    102 \\
\midrule
\textbf{Total}    &              &\textbf{155{,}506} & \textbf{24{,}355} & \textbf{17{,}328} & \textbf{5{,}464} \\
\bottomrule
\end{tabular}
\caption{\textbf{Per-source training-row counts across pipeline stages.}
\textsc{Raw} is the aggregated training pool from $14$ corpora;
\textsc{Diff.}\ is the survivors after the rule-based filters, NER grounding, and the MiniCheck difficulty band (\S\ref{sec:data:filter:rule}--\S\ref{sec:data:filter:diff});
\textsc{Dedup} adds semantic deduplication, test-set decontamination, and the $\geq{}2$-questions filter (\S\ref{sec:data:filter:dedup}--\S\ref{sec:data:filter:decomp});
\textsc{Final} is the curated $5{,}464$-claim training set after submodular diversity selection and long-evidence augmentation (\S\ref{sec:data:filter:diversity}--\S\ref{sec:data:filter:aug}).
The single largest cut is the difficulty band, which removes claims that strong fact-checkers already verify confidently as well as likely-noisy items at the low-confidence tail.}
\label{tab:source_distribution}
\end{table}

\section{Implementation Details}
\label{sec:appendix:impl}

\paragraph{LoRA configuration.}
We attach LoRA adapters~\citep{hu2022lora} of rank $64$ and $\alpha{=}128$ to all attention and MLP projections (\texttt{q,k,v,o,gate,up,down}); the base Qwen2.5-Instruct policy remains frozen in bfloat16.

\paragraph{Judge and embedding services.}
The Qwen3-32B judge is served via vLLM \citep{kwon2023vllm} at temperature $0$ with seed $42$ and a $4$k--$8$k token budget.
The embedding model for the diversity reward is Qwen3-Embedding-8B.
All judge prompts are deterministic, and responses are cached on-disk keyed by SHA-256 of the prompt and configuration, so the same trace structure receives the same reward across epochs and resumes.

\paragraph{Optimization.}
We use AdamW with learning rate $5{\times}10^{-6}$ on a cosine-with-min-LR schedule (min $5{\times}10^{-7}$), warmup ratio $0.1$, weight decay $0.001$, and gradient clipping at $1.0$.
Per-device batch size is $4$ and gradient accumulation brings the effective global batch to $16$.

\paragraph{GRPO specifics.}
We sample $G{=}8$ rollouts per prompt, use clipping $\epsilon{=}0.2$ with $\epsilon_\text{high}{=}0.28$~\citep{deepseek2025r1}, the BNPO loss variant, and no KL penalty ($\beta{=}0$).
We enable \texttt{mask\_truncated\_completions} to avoid rewarding length-cutoff degeneracies.

\section{Evaluation Benchmarks}
\label{sec:appendix:benchmarks}

We evaluate on $11$ held-out claim-verification benchmarks, partitioned into $9$ in-domain datasets that share at least one source corpus with the training pool and $2$ out-of-domain datasets used only at test time.
All datasets are cast as $2$-way (\supported{}/\refuted{}) tasks.

\paragraph{In-domain (9).}
The in-domain set spans three broad sources:
\textit{Wikipedia}: FEVER~\citep{thorne2018fever}, HoVer~\citep{ho2020constructing}, FEVEROUS~\citep{FactExtractionVERification_2021a}, WiCE~\citep{WiCERealWorldEntailment_2023b}, Ex-FEVER~\citep{EXFEVERDatasetMultihop_2024a}, and FoolMeTwice~\citep{FoolMeTwice_2021};
\textit{political claims}: ClaimDecomp~\citep{chen2022generating};
and \textit{biomedical or public-health text}: PubHealth~\citep{kotonya2020explainable} and PubMedClaim~\citep{PubMedQADatasetBiomedical_2019}.

\paragraph{Out-of-domain (2).}
CoverBench~\citep{jacovi2024coverbench} targets long-evidence multi-hop verification, and LLM-AggreFact~\citep{tang2024minicheck} aggregates factuality judgments across heterogeneous generators.
Neither corpus contributes to training, and both are used only to probe generalization beyond the training distribution.

\section{Result Plots}
\label{sec:appendix:scale}

\begin{figure*}[!h]
    \centering
    \includegraphics[width=0.9\textwidth]{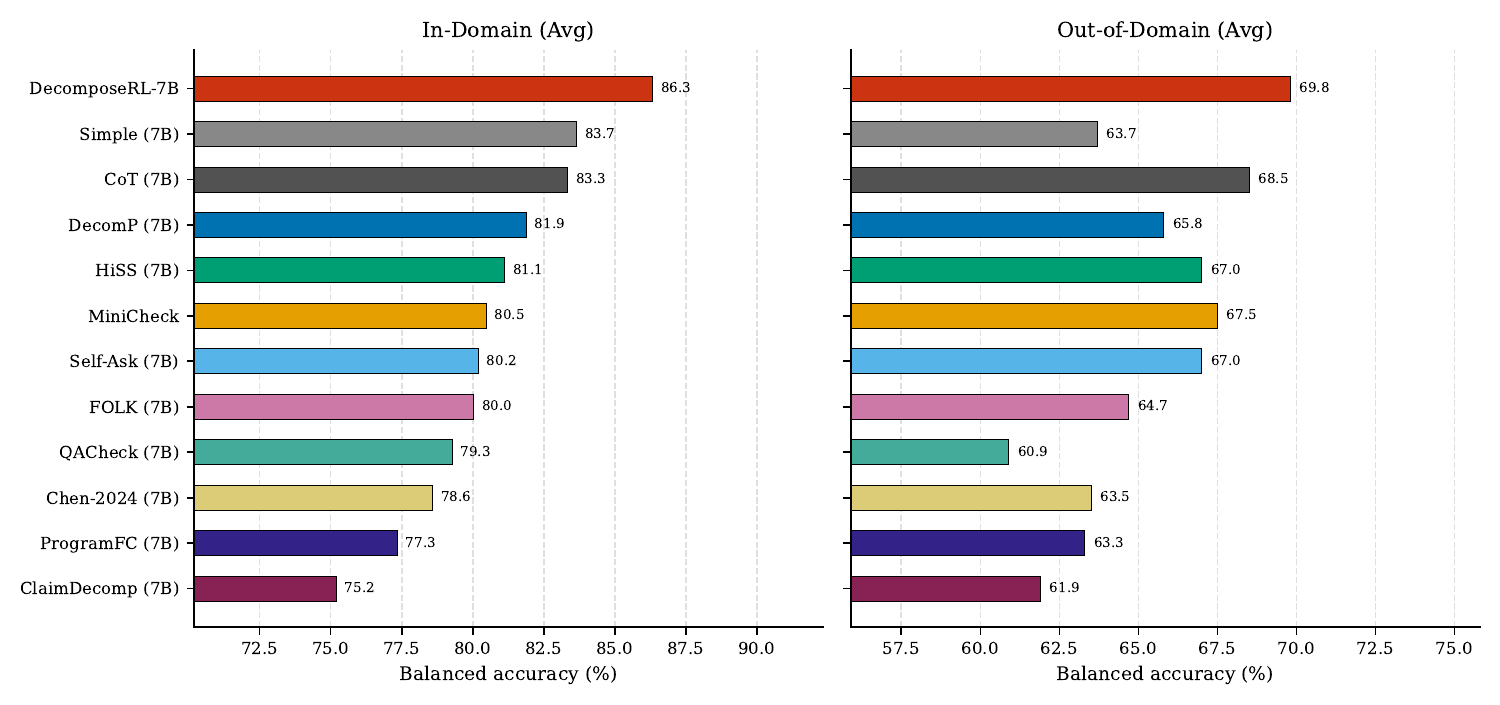}
    \caption{\textbf{Results comparing \method{} with multiple baselines.}
    Balanced accuracy (\%) of \method{} against $11$ baselines at matched scale, split into the in-domain Avg over $9$ datasets (left) and the out-of-domain Avg over CoverBench and LLM-AggreFact (right).
    \method{} (red) is the only system dominating every prompted baseline on both panels simultaneously.}
    \label{fig:main_results}
\end{figure*}

\begin{figure*}[!htbp]
    \centering
    \includegraphics[width=\linewidth]{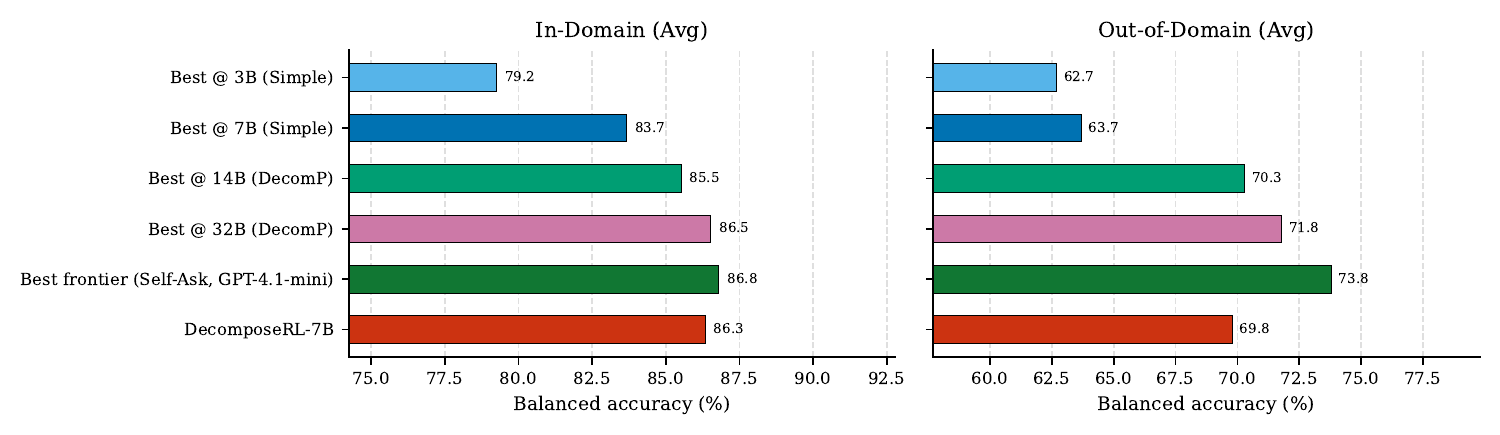}
    \caption{\textbf{\method{} vs.\ the best baseline at each scale.}
    In-domain Avg (left) and out-of-domain Avg (right) balanced accuracy for the strongest prompted baseline at each scale ($3$B, $7$B, $14$B, $32$B) and a proprietary frontier baseline.
    \method{} at $7$B matches the $32$B baseline and frontier on in-domain Avg and trails the larger baselines by less than $4$ points on out-of-domain Avg.}
    \label{fig:scale_comparison}
\end{figure*}

We complement the per-dataset numeric tables in \cref{sec:exp:main} with two bar-chart views of the headline result.
\cref{fig:main_results} shows the full $7$B comparison: \method{} against all $11$ baselines on the in-domain (Avg) and out-of-domain (Avg) aggregates, the same row set as \cref{tab:main_combined_7b}.
\cref{fig:scale_comparison} places \method{}-$7$B on a side-by-side view against the best-performing baseline at each parameter scale ($3$B, $7$B, $14$B, $32$B) plus the GPT-$4.1$-mini frontier, mirroring \cref{tab:main_combined_best}.
\method{} -- a $7$B policy trained with our reward ensemble -- attains in-domain Avg comparable to baselines that are $4{\times}$ to an order of magnitude larger; the clearest remaining gap is on the out-of-domain panel, where \method{} ($69.8$) trails Decomposed~Prompting~@~$14$B ($70.3$) by $0.5$ points despite using half the parameters, suggesting that the residual headroom is closable with additional supervision rather than additional parameters.

\section{Ablations} \label{app:ablations}

This section groups the controlled ablations that vary one axis at a time while holding the rest of the \method{} method fixed. 

\subsection{Diversity Selector Ablation}
\label{app:selector_comparison}

This section demonstrates the empirical claim made in \cref{sec:data:filter:diversity}: under the same per-source and per-label quotas, Facility-Location's max-similarity credit gives better coverage of the pool than greedy distance-maximizing selectors and is calibrated against the pool's outlier shell rather than being pulled into it.

\paragraph{Setup.}
We hold the post-dedup pool ($N{=}17{,}328$), the budget ($|S|{=}5{,}000$), the $50/50$ label balance, and the per-source $\sqrt{n_s}$ quotas fixed, and swap only the intra-cell selector among three families:
(i) \textbf{Submodular}: lazy-greedy Facility-Location on cosine similarity, the selector used by \method{};
(ii) \textbf{KMeans}: $k$-means with $k$ equal to the cell budget, returning the pool claim closest to each centroid;
(iii) \textbf{Farthest-Point}: a greedy $k$-center / MaxMin schedule that iteratively adds the pool claim whose minimum cosine distance to the already-selected set is largest.

\paragraph{Metrics.}
We report two coverage statistics and one outlier statistic, each derived directly from the same \texttt{text-embedding-3-large} cache used by \method{}:
\textbf{Cov.\,\%} is the share of non-empty $50{\times}50$ PCA bins of the pool that contain at least one selected claim.
\textbf{$d_{\mathrm{med}}$} and \textbf{$d_{95\%}$} are the median and $95$th-percentile cosine distance from a pool claim to its \emph{nearest} selected claim, computed on a fixed $3{,}000$-claim subsample of the pool; lower means a tighter covering radius and so better worst-case coverage. \textbf{Outlier\,\%} is the share of the selected set that lies in the pool's top-$5\%$ most isolated claims, where isolation is measured by mean cosine distance to a claim's $10$ in-pool nearest neighbors. A uniformly random sample under the same quotas would yield $5\%$; values above this indicate that the selector \emph{prefers} the outlier shell over the dense interior.

\begin{figure*}[!htbp]
    \centering
    \includegraphics[width=0.7\linewidth]{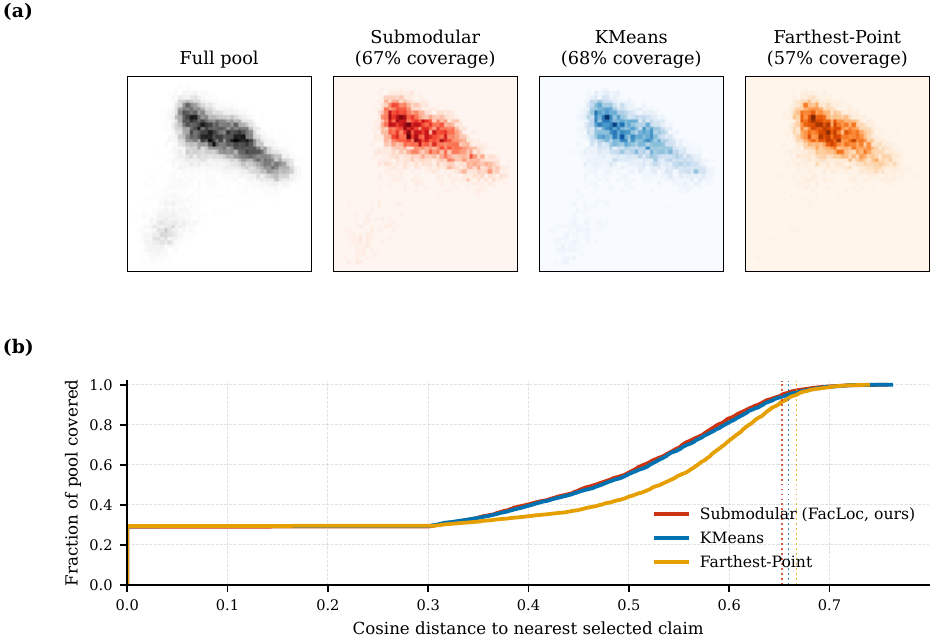}
    \caption{\textbf{Selector ablation on the post-dedup pool} ($N{=}17{,}328$ claims, budget $|S|{=}5{,}000$).
    \textbf{(a)} PCA density of the full pool (grey) and of each selector's pick (red / blue / orange), with coverage of non-empty pool bins annotated.
    \textbf{(b)} Cumulative fraction of the pool whose nearest selected claim lies within cosine distance $d$; lower-and-leftward is better worst-case coverage. Dotted verticals mark each method's $d_{95\%}$.
    Submodular and KMeans track each other closely; Farthest-Point gives up \textasciitilde$10$ percentage points of coverage to chase isolated points, consistent with the outlier-pulling failure mode discussed in \cref{sec:data:filter:diversity}.}
    \label{fig:selector_comparison}
\end{figure*}

\paragraph{Results.}
\cref{fig:selector_comparison} and \cref{tab:selector_comparison} report the comparison.
Submodular and KMeans cover the pool almost identically ($66.6\%$ vs.\ $68.3\%$ of populated bins; $d_{\mathrm{med}}$ within $0.006$), and both stay close to the $5\%$ baseline on outlier picks ($7.6\%$ and $7.0\%$). Farthest-Point exhibits the failure mode anticipated in \cref{sec:data:filter:diversity}: it concedes ${\sim}10$ percentage points of bin-level coverage and ${\sim}0.06$ on the median NN-distance to over-represent the outlier shell at $12.5\%$, i.e.\ ${\sim}1.6{\times}$ the rate of Submodular and ${\sim}2.5{\times}$ a uniform sample. The PCA panels in \cref{fig:selector_comparison}(a) make this geometric: the small detached cluster at the lower-left of the pool (visible in the grey panel) is up-weighted under Farthest-Point's MaxMin objective but is selected at most once per call by Submodular's $\max$-similarity credit. The CDF in panel~(b) is the integrated version of the same statement: Submodular and KMeans dominate Farthest-Point at every quantile of the NN-distance.

\paragraph{Cost.}
On the largest stratified cell ($n{=}3{,}482$, budget $430$, which dominates pipeline wall-clock), Facility-Location runs in $23.7$\,s versus $103.2$\,s for KMeans (the cost of running $k$-means with $k$ equal to the cell budget); Farthest-Point is essentially free at $0.1$\,s but pays for it in coverage. We adopt Submodular because it is the only selector in the table that is simultaneously better at coverage front and cheap enough to rerun under the per-source/per-label cell sweep of \cref{sec:data:filter:diversity}. Running a downstream-reward sweep over all three selectors is prohibitively costly under our GRPO training budget (each $5{,}000$-claim run takes ${\sim}48$\,h on $4$\,GPUs with multiple LLM-as-a-judge calls per rollout), so we select the diversity selector on this low-cost structural analysis rather than on end-task accuracy.

\begin{table*}[t]
  \centering
  \small
  \setlength{\tabcolsep}{4.5pt}
  \begin{tabular}{lrrrrrrr}
    \toprule
    Selector & $|S|$ & Cov.\,\% $\uparrow$ & $d_{\mathrm{med}}$ $\downarrow$ & $d_{95\%}$ $\downarrow$ & Outlier\,\% $\downarrow$ & $H_{\mathrm{src}}$ (bits) $\uparrow$ & $t$ (s) $\downarrow$ \\
    \midrule
  \textbf{Submodular (FacLoc, ours)} & 5000 & 66.6 & 0.467 & 0.653 & 7.6 & 3.28 & 23.7 \\
  KMeans & 5000 & 68.3 & 0.473 & 0.659 & 7.0 & 3.28 & 103.2 \\
  Farthest-Point & 5000 & 56.6 & 0.531 & 0.667 & 12.5 & 3.28 & 0.1 \\
    \bottomrule
  \end{tabular}
  \caption{\textbf{Selector ablation.} Submodular matches KMeans on pool coverage at $4{\times}$ lower kernel cost and picks $1.6{\times}$ fewer outliers than Farthest-Point. See \cref{app:selector_comparison} for metric definitions.}
  \label{tab:selector_comparison}
\end{table*}

\subsection{Sampling Ablation} \label{app:sampling_ablation}

\cref{app:selector_comparison} establishes at the \emph{pool} level that submodular Facility-Location wins on coverage and outlier-pick rate. This section ablates the strategy by checking whether the pool-level advantage carries through to \emph{downstream} accuracy: \textit{does the choice of sampler matter once the policy is trained on the resulting set?}

\paragraph{Setup.}
Two GRPO runs are compared, identical in every aspect except the sampler. The default run (\method{}-$7$B) draws the $5{,}464$ claims via submodular maximization of a Facility-Location objective with lazy-greedy (\cref{sec:data}); the contrastive run draws the same $5{,}464$ claims uniformly at random from the post-dedup pool.

\paragraph{Submodular sampling helps where coverage matters.}
\cref{tab:sampling_ablation} reports the comparison. Random sampling loses $1.5$ Avg on the in-domain suite ($86.3 \to 84.8$) and $4.3$ points on the out-of-domain CoverBench ($62.5 \to 58.2$). The gap is concentrated on the harder benchmarks rather than spread uniformly: LLM-AggreFact actually nudges up by $0.4$ under random sampling, which is inside the noise band we observe across reward-shuffle seeds. The asymmetry is consistent with the picture from \cref{app:selector_comparison}: submodular FacLoc gives up nothing on dense regions of the embedding pool but covers the long-evidence and tabular tail that CoverBench probes; uniform sampling under-represents that tail in proportion to its mass in the pool. The reward stack alone is therefore not sufficient: feeding it a representative training set is what unlocks the performance gain.

\begin{table*}[t]
\centering
\small
\setlength{\tabcolsep}{4pt}
\renewcommand{\arraystretch}{1.05}
\begin{tabular}{l c c c c}
\toprule
 & & \textbf{In-Domain} & \multicolumn{2}{c}{\textbf{Out-of-Domain}} \\
\cmidrule(lr){3-3} \cmidrule(lr){4-5}
\textbf{Variant} & \textbf{Sampling} & \textbf{Avg} & \textbf{CoverBench} & \textbf{LLM-AggreFact} \\
\midrule
\method{}-$7$B (default)         & Submodular (FacLoc) & \textbf{86.3} & \textbf{62.5} & 77.0 \\
\method{}-$7$B (random sampling) & Random i.i.d.       & 84.8 & 58.2 & \textbf{77.4} \\
\addlinespace[2pt]
\rowcolor{gray!10}
$\Delta$ (random $-$ submodular) & $-$ & $-1.5$ & $-4.3$ & $+0.4$ \\
\bottomrule
\end{tabular}
\caption{\textbf{Submodular sampling beats random sampling at the same budget.}
Same $5{,}464$-claim training set size, same reward stack, same hyperparameters; only the strategy for drawing the $5{,}464$ claims from the post-dedup pool ($N{=}17{,}328$, see \cref{app:selector_comparison}) changes. Submodular Facility-Location with lazy-greedy gives $+1.5$ in-domain Avg and $+4.3$ on the harder out-of-domain CoverBench; the gain is concentrated on the long-evidence tail rather than the in-distribution suite. LLM-AggreFact moves the opposite way ($-0.4$) but stays inside the per-run noise band.}
\label{tab:sampling_ablation}
\end{table*}

\subsection{Supervision-Rate Sweep} \label{app:supervision}

\cref{sec:exp:semisup} reports the two endpoints of the supervision sweep ($s{=}0.1$ vs.\ $s{=}1.0$); this section shows the full curve over $s\in\{0.1, 0.3, 0.5, 0.7, 1.0\}$.

\paragraph{Setup.}
Five GRPO runs, identical except for the supervision rate $s$ (\cref{sec:method:selfsup}). The labeled / unlabeled partition is computed once per claim per run so the assignment is consistent across epochs.

\begin{figure*}[t]
    \centering
    \includegraphics[width=\linewidth]{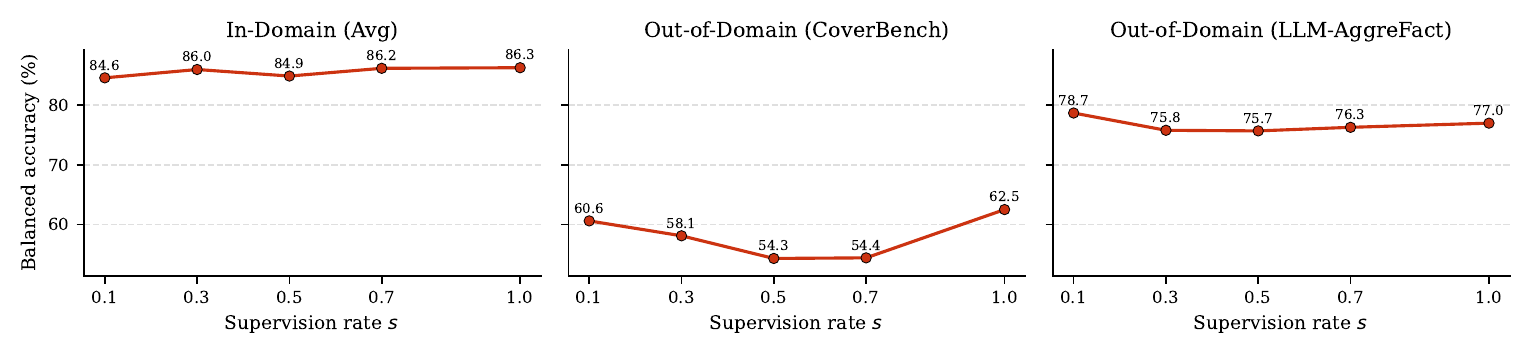}
    \caption{\textbf{Supervision-rate sweep.}
    \method{}-$7$B accuracy as a function of the supervision rate $s$ (fraction of training claims with a ground truth label; the remaining $1{-}s$ use the self-consistency and relative-necessity fallbacks from \cref{sec:method:selfsup}).
    In-domain Avg (left) is essentially flat -- the policy can be trained with as little as $10\%$ verdict supervision at a $1.7$-point cost.
    LLM-AggreFact (right) is also flat: standard-length claim verification does not need the ground truth label either.
    The only U-shaped axis is CoverBench (middle), where intermediate $s$ trails the endpoints by \textasciitilde$8$ points.
    We attribute this to mixed-signal interference between the verdict-conditioned and label-free reward branches and discuss it in \cref{app:supervision}.}
    \label{fig:supervision_curve}
\end{figure*}

\paragraph{In-domain and LLM-AggreFact are flat across the sweep.}
\cref{tab:supervision_full} and \cref{fig:supervision_curve} report the full numbers. In-domain Avg stays within a $1.7$-point band over the entire sweep ($84.6{-}86.3$); LLM-AggreFact stays within $3.0$ points ($75.7{-}78.7$), with $s{=}0.1$ actually scoring highest. On both of these axes the reward stack's self-consistency coverage and relative-necessity components recover most of what is lost by dropping the verification reward and the gold-conditioned necessity scoring.

\paragraph{CoverBench is the exception.}
The only axis with a visible curve is CoverBench, which traces a U-shape: $62.5$ at $s{=}1.0$, drops to $54.3{-}54.4$ at $s\in\{0.5, 0.7\}$, and recovers to $60.6$ at $s{=}0.1$. 
CoverBench has the most long-evidence reasoning, where the policy has the least training-set support already even in full supervision -- only $8.5\%$ of the training data (see \cref{sec:data:filter:aug}, \cref{fig:data_funnel}).
At intermediate $s\in\{0.3, 0.5, 0.7\}$, that ${\sim}8.5\%$ slice has to split its gradient between the verdict-conditioned reward branch (verification, gold-coverage, gold-necessity) and its label-free counterpart (self-consistency coverage, relative necessity), and neither branch sees enough long-evidence examples to converge on a consistent update direction.
Standard-length verification (LLM-AggreFact, all $9$ in-domain benchmarks) has roughly $10{\times}$ the training support and is robust enough to absorb the same mixed signal -- consistent with its essentially flat curve in \cref{fig:supervision_curve}.
At $s{=}1.0$ the full $8.5\%$ long-evidence slice receives the verdict-conditioned reward and CoverBench is trained on directly. At $s{=}0.1$ that supervised signal is effectively gone -- only a negligible ${\sim}0.8\%$ of the training pool ($\approx{10\%}\times{8.5\%}$) is both long-evidence and labeled -- and the policy instead leans on \texttt{Qwen2.5-7B-Instruct}'s pretrained long-context capability to handle CoverBench at inference time. The intermediate rates have neither pathway intact: the supervised long-evidence signal is partial and the policy is still being actively reshaped, so it cannot cleanly default to pretraining either.

A multi-seed sweep would tighten the confidence band around the observed U-shape, but the computational cost of GRPO training makes this prohibitive; we leave it to future work.

\begin{table*}[t]
\centering
\small
\setlength{\tabcolsep}{4pt}
\renewcommand{\arraystretch}{1.05}
\resizebox{\textwidth}{!}{%
\begin{tabular}{@{}c@{\hspace{6pt}}ccccccccc@{\hspace{8pt}}c@{\hspace{8pt}}ccc@{}}
\toprule
 & \multicolumn{9}{c}{\textbf{In-Domain (9 datasets)}} & & \multicolumn{3}{c}{\textbf{Out-of-Domain (2 datasets)}} \\
\cmidrule(lr){2-11} \cmidrule(lr){12-14}
\textbf{$s$} & \textbf{FEVER} & \textbf{ClaimDecomp} & \textbf{HoVer} & \textbf{FEVEROUS} & \textbf{WiCE} & \textbf{Ex-FEVER} & \textbf{PubHealth} & \textbf{PubMedClaim} & \textbf{FoolMeTwice} & \textbf{Avg} & \textbf{CoverBench} & \textbf{LLM-AggreFact} & \textbf{Avg} \\
\midrule
$0.1$ & 71.4 & 98.1 & 70.4 & 92.9 & 87.9 & 82.6 & 86.9 & 83.9 & 87.1 & 84.6 & \underline{60.6} & \textbf{78.7} & \underline{69.7} \\
$0.3$ & 73.3 & \textbf{100.0} & 72.2 & 93.1 & \underline{89.5} & 83.7 & \textbf{89.9} & 85.4 & 86.8 & 86.0 & 58.1 & 75.8 & 67.0 \\
$0.5$ & 72.1 & 98.6 & 70.8 & \underline{93.9} & \textbf{90.1} & 82.3 & 86.9 & 83.9 & 85.4 & 84.9 & 54.3 & 75.7 & 65.0 \\
$0.7$ & \textbf{74.2} & \textbf{100.0} & 74.6 & \textbf{94.3} & 87.7 & 85.9 & 87.2 & 85.1 & 87.2 & \underline{86.2} & 54.4 & 76.3 & 65.4 \\
$1.0$ & \underline{74.1} & 98.6 & \textbf{76.4} & 93.1 & 86.5 & \textbf{87.6} & \underline{87.5} & \textbf{85.5} & \textbf{87.7} & \textbf{86.3} & \textbf{62.5} & \underline{77.0} & \textbf{69.8} \\
\bottomrule
\end{tabular}}
\caption{\textbf{Full supervision-rate sweep.} All rows are \method{}-$7$B, varying only the supervision rate $s$ (fraction of training claims with a ground truth verdict; the rest use pseudo-label fallbacks, \cref{sec:method:selfsup}). In-domain Avg is essentially flat ($84.6{-}86.3$); LLM-AggreFact varies within $3.0$ points; the only U-shaped axis is CoverBench ($54.3{-}62.5$), discussed in \cref{app:supervision}. \textbf{Bold}: column-best; \underline{underline}: runner-up.}
\label{tab:supervision_full}
\end{table*}

\subsection{Reward Ensemble Ablation} \label{app:full_reward_ablation}

\cref{sec:exp:reward} reports the compact leave-one-out ablation; this section provides the full per-dataset breakdown and a visual summary.

\paragraph{Setup.}
Five GRPO runs, each identical to the default \method{}-$7$B run except that one reward is removed. Format ($R_\text{fmt}$) and Verification ($R_\text{ver}$) are excluded from ablation: they are structural prerequisites whose removal causes training collapse (the policy loses its output-shape constraint).

\paragraph{Per-dataset breakdown.}
\cref{tab:reward_ablation_full} reports the full $11$-dataset comparison. The per-dataset scores fluctuate by $2{-}3$ points in either direction, but the in-domain Avg stays within $0.7$ of the full ensemble across all five removals -- the in-domain signal has enough redundancy to tolerate losing any single reward. Out-of-domain tells a different story: CoverBench drops by $1.6{-}8.6$ points depending on the reward removed, and LLM-AggreFact drops by $0.9{-}3.0$ points.

\paragraph{Necessity is the dominant OOD signal.}
Removing the leave-one-out Necessity reward (\cref{sec:method:necessity}) causes the largest single-reward CoverBench collapse: $62.5{\to}53.9$ ($-8.6$), more than double the next-largest drop (Coverage at $-5.4$). This is consistent with the reward's design: the $4$-state saliency matrix (necessary / redundant / neutral / harmful) forces every sub-question to earn its place by flipping the verdict when removed, which is exactly the kind of signal long-evidence, multi-hop reasoning needs. Without it, the policy can satisfy every other reward by generating plausible-looking sub-questions that happen not to cover the critical evidence chain.

\paragraph{Coverage and Diversity form the next tier.}
Coverage removal ($-3.9$ OOD Avg) and Diversity removal ($-3.2$) both degrade the policy's ability to span the claim: Coverage checks whether the answers collectively predict the correct verdict; Diversity penalises redundant sub-questions via MMR over embeddings. Together they enforce the ``breadth'' axis of decomposition quality, and their OOD drops are correspondingly larger than those of Joint Quality ($-2.0$) and Question Count ($-1.7$), which operate on ``depth'' (per-question correctness and trace length).

\begin{figure*}[t]
    \centering
    \includegraphics[width=\linewidth]{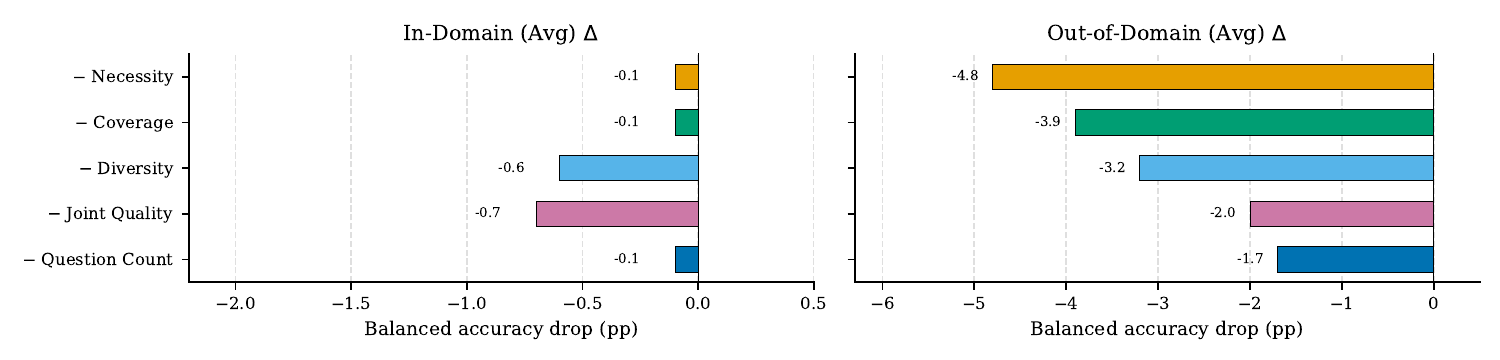}
    \caption{\textbf{Reward ablation} (same data as \cref{tab:reward_ablation_full}).
    In-domain Avg (left) is robust to removing any single reward ($\le 0.7$ pp drop).
    Out-of-domain Avg (right) reveals the contribution of each: Necessity ($-4.8$) and Coverage ($-3.9$) contribute the most, consistent with their role in measuring whether the decomposition collectively covers the claim.}
    \label{fig:reward_ablation}
\end{figure*}

\paragraph{Visual summary.}
\cref{fig:reward_ablation} plots the same in-domain and out-of-domain Avg drops side by side. The contrast between the two panels -- tiny bars on the left, large bars on the right -- is the clearest visual evidence that the reward ensemble is necessary for generalization rather than for in-distribution accuracy.

\begin{table*}[t]
\centering
\small
\setlength{\tabcolsep}{4pt}
\renewcommand{\arraystretch}{1.05}
\resizebox{\textwidth}{!}{%
\begin{tabular}{@{}l@{\hspace{6pt}}cccccccccc@{\hspace{8pt}}ccc@{}}
\toprule
 & \multicolumn{10}{c}{\textbf{In-Domain (9 datasets)}} & \multicolumn{3}{c}{\textbf{Out-of-Domain (2 datasets)}} \\
\cmidrule(lr){2-11} \cmidrule(lr){12-14}
\textbf{Variant} & \textbf{FEVER} & \textbf{ClaimDecomp} & \textbf{HoVer} & \textbf{FEVEROUS} & \textbf{WiCE} & \textbf{Ex-FEVER} & \textbf{PubHealth} & \textbf{PubMedClaim} & \textbf{FoolMeTwice} & \textbf{Avg} & \textbf{CoverBench} & \textbf{LLM-AggreFact} & \textbf{Avg} \\
\midrule
\textbf{\method{}} & \textbf{74.1} & 98.6 & \textbf{76.4} & 93.1 & 86.5 & \textbf{87.6} & 87.5 & 85.5 & 87.7 & \textbf{86.3} & \textbf{62.5} & \textbf{77.0} & \textbf{69.8} \\
\addlinespace[2pt]
$-$ Necessity      & 73.3 & 98.6 & 74.0 & \textbf{94.4} & \underline{89.0} & 87.3 & \underline{88.3} & 83.8 & 87.5 & \underline{86.2} & 53.9 & 76.1 & 65.0 \\
$-$ Coverage       & 72.8 & 98.6 & 73.4 & \underline{94.0} & \textbf{89.5} & 86.4 & 87.3 & \underline{86.2} & 87.3 & 86.2 & 57.1 & 74.6 & 65.9 \\
$-$ Diversity      & 71.9 & 97.3 & 73.6 & 93.6 & 89.0 & 86.6 & 86.9 & 85.6 & 86.5 & 85.7 & 58.2 & 74.9 & 66.6 \\
$-$ Joint Quality  & 72.7 & 97.3 & \underline{74.2} & 93.8 & 87.4 & 86.8 & 85.5 & \textbf{86.3} & 86.9 & 85.6 & \underline{61.5} & 74.0 & \underline{67.8} \\
$-$ Question Count & \underline{72.6} & \textbf{100.0} & 71.8 & 93.5 & 86.3 & 86.1 & \textbf{89.9} & 87.3 & \textbf{88.0} & 86.2 & 60.9 & \underline{75.3} & 68.1 \\
\bottomrule
\end{tabular}}
\caption{\textbf{Full per-dataset reward ablation.} Extends \cref{tab:reward_ablation} with per-dataset scores. Each row removes one reward while keeping everything else fixed. \textbf{Bold}: best within the ablation set; \underline{underline}: runner-up. Necessity removal causes the largest CoverBench collapse ($62.5{\to}53.9$, $-8.6$); Coverage and Diversity removals also degrade CoverBench substantially ($-5.4$ and $-4.3$). In-domain Avg stays within $0.7$ across all five removals.}
\label{tab:reward_ablation_full}
\end{table*}

\subsection{Model Scaling} \label{app:model_scaling}

The main paper reports a single $7$B policy (\cref{sec:exp:main}); this section isolates the contribution of \emph{the reward stack} from the contribution of \emph{policy size} by re-running \method{} on a smaller \texttt{Qwen2.5-3B-Instruct} base and comparing it against every other method evaluated at the same parameter scale.

\paragraph{Setup.}
We swap only the base model and keep everything else identical to the main run. We refer to this run as \method{}-$3$B; the $7$B run from the main paper is \method{}-$7$B. The $3$B comparison pool is every other method evaluated with a $3$B base under the same protocol: the two non-decomposition base prompts (Simple, CoT) and the eight prompted decomposers (Self-Ask, Decomposed Prompting, HiSS, FOLK, ProgramFC, Chen-$2024$, ClaimDecomp, QACheck).

\begin{figure*}[t]
    \centering
    \includegraphics[width=\linewidth]{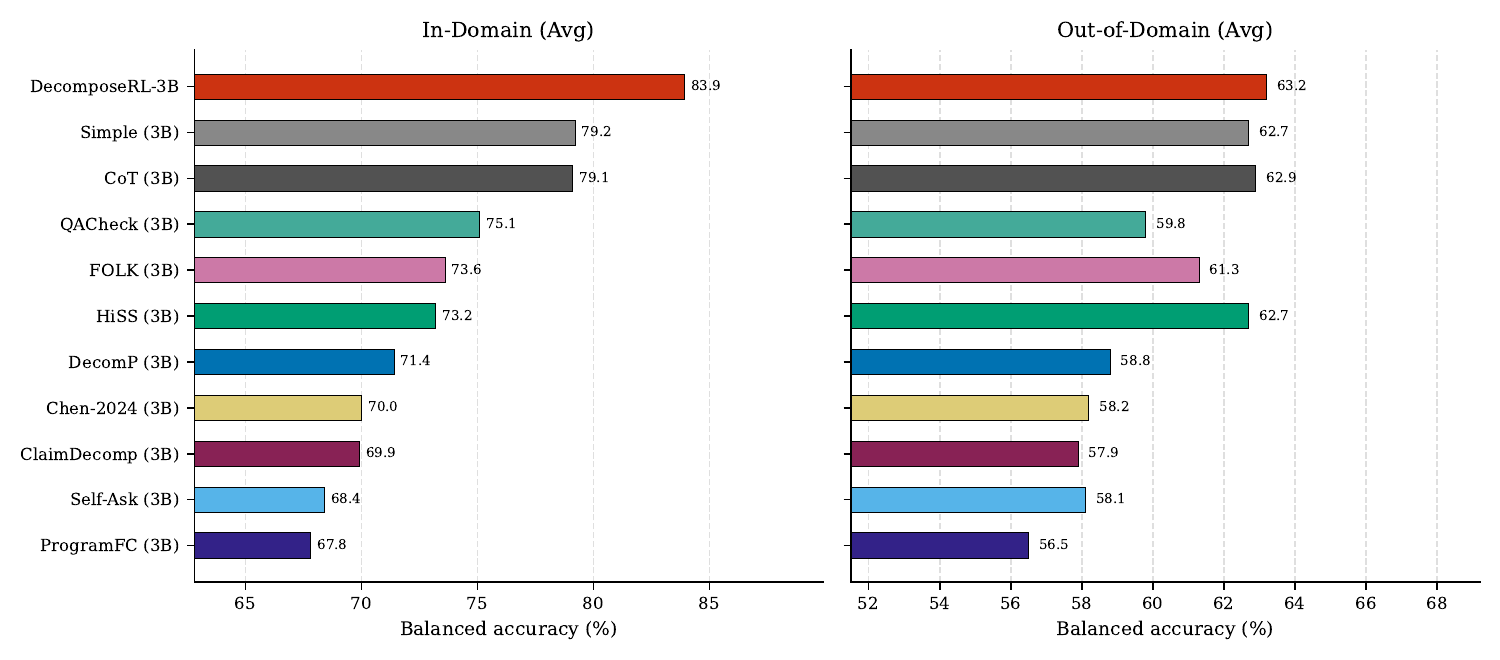}
    \caption{\textbf{3B ablation.}
    All methods evaluated with a Qwen-$3$B base on the same in-domain ($9$ datasets) and out-of-domain ($2$ datasets) suite.
    \method{}-$3$B is the only $3$B configuration that simultaneously tops both panels, beating the strongest $3$B baseline by $+4.7$ on in-domain Avg and $+0.6$ on out-of-domain Avg.
    The reward stack therefore transfers to a smaller policy without retuning, isolating the gain from policy size.}
    \label{fig:model_scaling}
\end{figure*}

\paragraph{The reward ensemble still wins at 3B.}
\cref{fig:model_scaling} reports the full $3$B comparison. \method{}-$3$B reaches $83.9$ in-domain Avg and $63.2$ out-of-domain Avg, the highest score in both panels.
The gap against the strongest non-decomposition $3$B baseline (Simple at $79.2$ / $62.7$) is $+4.7$ Avg and $+0.5$ OOD; the gap against the strongest $3$B prompted decomposer (QACheck at $75.1$ / $59.8$) is $+8.8$ Avg and $+3.4$ OOD. Because the only variable changed within this panel is the training objective (reward-ensemble RL vs.\ prompting/imitation), the gain is attributable to the reward ensemble rather than to policy capacity.

\paragraph{Scaling the policy keeps paying off.}
Comparing this appendix figure to the $7$B main result (\cref{tab:main_combined_7b}, $86.3$ Avg / $69.8$ OOD), going from \method{}-$3$B to \method{}-$7$B adds $+2.4$ in-domain Avg and $+6.6$ out-of-domain Avg without any change to the training recipe. The larger of the two deltas is on the out-of-domain side, consistent with the intuition that long-evidence, multi-hop generalization is more parameter-bound than in-distribution verification: the $7$B policy has more capacity to keep multiple sub-claims coherent across a long context window. Due to compute constraints, we did not train \method{} at $14$B or $32$B scale. Nevertheless, the fact that \method{}-$7$B already approaches GPT-$4.1$-mini frontier performance ($86.3$ vs.\ $86.8$ in-domain average at one-to-two orders of magnitude fewer parameters; \cref{tab:main_combined_best}) suggests the scaling curve is not yet saturated, and we leave a more thorough exploration of policy-side scaling to future work.

\subsection{Tiny-Judge Ablation} \label{app:tiny_judge}

Five of \method{}'s seven rewards are scored by a \texttt{Qwen3-32B} LLM judge (\cref{sec:method:rewards}), and that judge dominates the training-time GPU footprint. This section runs the controlled experiment: same data, same reward stack, same policy, same training schedule -- only the judge changes. We compare the default LLM-judge run against a ``tiny-judge'' variant that replaces the $32$B judge with eight task-specific LoRA adapters fine-tuned on a shared \texttt{ModernBERT-large} backbone, an ${\sim}10{\times}$ reduction in judge parameter count ($32$B $\to$ ${\sim}400\text{M} \times 8 = 3.2$B total).

\paragraph{Setup.}
The tiny judge is a stack of eight LoRA-adapted classifier heads on a shared ModernBERT-large encoder (${\sim}400$M parameters), one head per judge task (the five atomicity sub-criteria, answerability, answer correctness, and coverage; the necessity reward re-uses the coverage head). Each head is trained off-line on labels distilled from earlier Qwen$3$-$32$B judge calls, then frozen during the GRPO training of the policy.\footnote{We will release the distilled training traces and the trained tiny-judge LoRA checkpoints upon acceptance.} The training pools range from ${\sim}1.2$k to ${\sim}250$k labelled examples per task (small heads only need a few thousand; the coverage head consumes the larger pool), and training all eight heads end-to-end costs ${\sim}24$ H$100$-hours -- a one-time upfront expense that we discuss in the compute analysis below. Both runs are evaluated on the full $9$ in-domain and $2$ out-of-domain benchmarks (\cref{sec:appendix:benchmarks}).

\paragraph{Compute saving is large; accuracy loss is small.}
\cref{tab:tiny_judge_ablation} reports the comparison. The default LLM-judge run requires $120$\,h of wall-clock on a $2\times$H$100$ setup (one GPU for training, one GPU hosting the $32$B judge via vLLM); the tiny-judge run finishes in $48$\,h on a single H$100$ that hosts both training and reward scoring. The judge stack shrinks from $32$B to ${\sim}3.2$B parameters ($8\times 400$M, ${\sim}10{\times}$ smaller), total compute drops from $240$ to $48$ GPU-hours ($-80\%$), and wall-clock drops by $60\%$. The accuracy cost on the matched evaluation suite is $-0.5$ in-domain Avg ($86.3{\to}85.8$) and roughly one point on each of the two out-of-domain benchmarks. The tradeoff is therefore favorable: a deployer recovers ${\sim}99\%$ of the LLM-judge in-domain Avg at one-fifth the GPU-hours and an order-of-magnitude smaller judge.

\paragraph{The one-time judge-training cost amortizes quickly.} Training the \emph{tiny judge} incurs a fixed overhead of ${\sim}24$ H$100$-hours, which is added to the first GRPO run but paid only once across all subsequent runs. On a single run, the total cost is $24$h (judge training) $+$ $48$h (GRPO) $= 72$h, compared to $240$h for the LLM-judge baseline -- a $70$\% reduction. Crucially, this advantage compounds with scale: over $K$ runs, the total cost is $240K$ hours for the LLM-judge versus $24 + 48K$ hours for ``tiny-judge'', making the relative saving strictly increasing in $K$.

\begin{table*}[t]
\centering
\small
\setlength{\tabcolsep}{4pt}
\renewcommand{\arraystretch}{1.05}
\resizebox{\textwidth}{!}{%
\begin{tabular}{l c c c c c c c}
\toprule
 & & & & & \textbf{In-Domain} & \multicolumn{2}{c}{\textbf{Out-of-Domain}} \\
\cmidrule(lr){6-6} \cmidrule(lr){7-8}
\textbf{Variant} & \textbf{Judge params} & \textbf{Wall-clock} & \textbf{GPUs} & \textbf{GPU-hr} & \textbf{Avg} & \textbf{CoverBench} & \textbf{LLM-AggreFact} \\
\midrule
\method{}-$7$B (LLM judge, default) & $32$B (single model)      & $120$\,h ($5$\,d) & $2{\times}$H100 & $240$ & \textbf{86.3} & \textbf{62.5} & \textbf{77.0} \\
\method{}-$7$B (tiny judge)         & $8\times 400$M (\,${\approx}3.2$B\,)  & $48$\,h  ($2$\,d) & $1{\times}$H100 & $48$  & 85.8          & 61.0          & 76.4 \\
\addlinespace[2pt]
\rowcolor{gray!10}
$\Delta$ (tiny $-$ LLM) & ${\sim}10{\times}$ smaller & $-60\%$ & $-50\%$ & $-80\%$ & $-0.5$ & $-1.5$ & $-0.6$ \\
\bottomrule
\end{tabular}}
\caption{Replacing the Qwen$3$-$32$B LLM judge with eight LoRA-fine-tuned ModernBERT-large classifiers (the ``tiny judge'') cuts judge size by ${\sim}10{\times}$ and total GPU-hours by $80\%$ at the cost of $<1$ in-domain Avg point.}
\label{tab:tiny_judge_ablation}
\end{table*}

\subsection{Compute Scaling} \label{app:compute_scale}

A natural question is whether the second epoch over the curated $5{,}464$-claim pool is necessary. This section runs the controlled experiment: same data, same reward stack, same hyperparameters, only the epoch count varies.

\paragraph{Setup.}
We compare the default training schedule used throughout the main paper ($2$ epochs over the curated pool, $5{,}464$ optimizer steps) against a half-budget variant ($1$ epoch, $2{,}732$ optimizer steps). Everything else is held constant
Both runs are scored on the full $9$ in-domain and $2$ out-of-domain benchmarks (\cref{sec:appendix:benchmarks}).

\paragraph{In-domain accuracy is essentially flat.}
\cref{tab:epoch_ablation} reports the comparison. The half-budget run loses only $0.1$ Avg on the $9$-dataset in-domain suite. To primarily target the in-domain benchmarks the second epoch contributes essentially nothing, and the curated pool is small enough that a $1$-epoch run finishes in roughly half the wall-clock.

\paragraph{Out-of-domain accuracy needs the second epoch.}
The picture changes on the two held-out OOD benchmarks: CoverBench drops by $3.9$ points ($62.5{\to}58.6$) and LLM-AggreFact by $4.8$ points ($77.0{\to}72.2$) at the half budget. Connecting this with the model-scaling section (\cref{app:model_scaling}), which found that the largest gains from going $3$B$\to$$7$B were also concentrated on OOD, suggests a consistent pattern: out-of-domain, long-evidence generalization is the part of \method{} that is hungry for both more parameters and more gradient updates, while in-domain verification is comparatively easy to satisfy. The second epoch is therefore not free padding; it is what closes the long-evidence / cross-domain gap.

\paragraph{Reducing Policy Size.} To isolate the gain from policy size, we also report a controlled $3$B ablation in \cref{app:model_scaling}: holding the reward ensemble and training data fixed. \method{}-$3$B still tops both panels against every same-size baseline ($+4.7$ Avg, $+0.5$ OOD over the strongest $3$B baseline).
\cref{app:compute_scale} similarly halves the training budget ($1$ epoch vs.\ $2$): in-domain Avg is preserved within noise ($-0.1$) while out-of-domain regresses by ${\sim}4$ points, locating the gain from the second epoch on the long-evidence tail rather than on the in-domain suite.

\begin{table*}[t]
\centering
\small
\setlength{\tabcolsep}{4pt}
\renewcommand{\arraystretch}{1.05}
\begin{tabular}{l c c c c c}
\toprule
 & & & \textbf{In-Domain} & \multicolumn{2}{c}{\textbf{Out-of-Domain}} \\
\cmidrule(lr){4-4} \cmidrule(lr){5-6}
\textbf{Variant} & \textbf{Epochs} & \textbf{Steps} & \textbf{Avg} & \textbf{CoverBench} & \textbf{LLM-AggreFact} \\
\midrule
\method{}-$7$B (default)     & $2$ & $5{,}464$ & \textbf{86.3} & \textbf{62.5} & \textbf{77.0} \\
\method{}-$7$B (half-budget) & $1$ & $2{,}732$ & 86.2 & 58.6 & 72.2 \\
\addlinespace[2pt]
\rowcolor{gray!10}
$\Delta$ (1\,ep $-$ 2\,ep) & $-1$ & $-50\%$ & $-0.1$ & $-3.9$ & $-4.8$ \\
\bottomrule
\end{tabular}
\caption{\textbf{Halving the training budget keeps in-domain accuracy but costs out-of-domain.}
Same $5{,}464$-claim curated training set, same reward stack, same hyperparameters; only the number of epochs changes (and therefore the optimizer step count). At half the gradient steps the in-domain Avg is statistically indistinguishable from the default run ($-0.1$), but both out-of-domain benchmarks drop by ${\sim}4$ points, suggesting the second epoch is doing most of its work on the long-evidence / cross-domain tail.}
\label{tab:epoch_ablation}
\end{table*}

\section{Qualitative Analysis} \label{app:qualitative}

This section collects representative inputs to and outputs from \method{} for inspection: a sample claim from the curated training set with its silver-decomposition (\cref{app:data_example}), and four trained-policy verification traces spanning a multi-hop success, a clean refutation, a calibrated abstention, and a counting-style failure (\cref{app:trace_examples}).

\subsection{Training-Set Example}
\label{app:data_example}

\cref{fig:data_example} shows a representative claim from the curated training set with its silver decomposition.

\begin{figure}[!htbp]
\small
\hrule\vspace{2pt}
\textbf{Source.} LLM-AggreFact \quad \textbf{Label:} \refuted \quad $n^\star{=}3$
\vspace{2pt}\hrule\vspace{4pt}
\textbf{Claim.} ``Lil Jon's top ranked Billboard song was \emph{Get Low}.''

\textbf{Evidence (excerpt).} ``\emph{Get Low} is a song by American rap group Lil Jon \& the East Side Boyz, featuring American hip hop duo Ying Yang Twins, released as a single in 2003.\ldots\ \emph{Get Low} peaked at number two on the Billboard Hot 100\ldots''

\textbf{Ground Truth sub-questions.}
\begin{enumerate}
\setlength{\itemsep}{0pt}
\item Is \emph{Get Low} a song by Lil Jon \& the East Side Boyz (featuring the Ying Yang Twins)?
\item Does the document state that \emph{Get Low} peaked at number two on the Billboard Hot 100?
\item Does the document mention any Lil Jon song that achieved a Billboard Hot 100 peak higher than number two?
\end{enumerate}
\vspace{2pt}\hrule
\caption{\textbf{A representative training claim.} The silver decomposition (\cref{sec:data:filter:decomp}) isolates two factual checks (existence, Hot 100 peak) and a comparative check that pins down the ``top ranked'' qualifier.}
\label{fig:data_example}
\end{figure}

\subsection{Verification Traces} \label{app:trace_examples}

We first walk through the trace for the intro-teaser claim (\cref{fig:intro_teaser}) in \cref{fig:trace_orwell}, whose sub-questions and answers are generated using \method{} and evidence document are retrieved using off-the-shelf web retriever\footnote{\url{https://pypi.org/project/duckduckgo-search}} (the claim is not from any benchmark).
We then collect four representative \method{} traces from benchmark dumps, each showcasing a distinct aspect of the decomposition policy: a multi-hop composition success, a clean single-fact refutation, a calibrated abstention on a partially-unsupported claim, and a counting-style failure on an out-of-domain claim.
The four benchmark traces are drawn verbatim from the raw outputs on the held-out test split of the named benchmark; answers are lightly shortened for typesetting while preserving meaning.

\paragraph{Color legend.}
In each trace below, \textcolor{blue!55!black}{blue} marks evidence-grounded sub-questions and their answers, \textcolor{orange!70!black}{orange} marks calibrated abstention (``I don't know''), \textcolor{green!45!black}{green} marks a correct \method{} verdict and \textcolor{red!65!black}{red} marks an incorrect verdict.
Superscript markers ($\textsuperscript{1}$, $\textsuperscript{2}$, $\ldots$) link each answer to the labelled evidence chunk it draws from.

\begin{figure*}[t]
\centering
\begin{tcolorbox}[
    width=\textwidth,
    colback=white,
    colframe=black!55,
    boxrule=0.5pt,
    arc=2pt,
    left=6pt, right=6pt, top=5pt, bottom=5pt,
    fontupper=\small,
]
\textbf{Claim} \hfill \textcolor{black!55}{\footnotesize Intro-teaser claim (web-retrieved evidence)}\\[1pt]
\textit{``The author of Nineteen Eighty-Four and Animal Farm was awarded the Nobel Prize in Literature.''}

\vspace{4pt}\hrule height 0.3pt\vspace{4pt}

\textbf{Evidence}\\[1pt]
\textcolor{black!75}{%
\textsuperscript{1}\,\textbf{George Orwell} ($1903$--$1950$), born Eric Arthur Blair, was an English novelist and essayist whose principal works include \emph{Animal Farm} ($1945$) and \emph{Nineteen Eighty-Four} ($1949$), both published under the pen name Orwell.\,
\textsuperscript{2}\,\textbf{Honours and awards.} Orwell received the Prometheus Hall of Fame Award (posthumous, $2011$) for \emph{Nineteen Eighty-Four} and a Retrospective Hugo Award ($1996$) for \emph{Animal Farm}; he was not a recipient of the Nobel Prize in any category.}

\vspace{4pt}\hrule height 0.3pt\vspace{4pt}

\textbf{\method{} verification trace}\\[2pt]
\vspace{-10pt}
\begin{tcolorbox}[boxsep=0pt, left=4pt, right=4pt, top=2pt, bottom=2pt,
    colback=blue!4, colframe=blue!35, boxrule=0.3pt, arc=1.5pt]
\textcolor{blue!55!black}{\textbf{Q$_1$}:}\ Who wrote \emph{Nineteen Eighty-Four}?\\
\textcolor{blue!55!black}{\textbf{A$_1$}:}\ George Orwell wrote \emph{Nineteen Eighty-Four}, published in $1949$.\,\textcolor{black!55}{\tiny\textsuperscript{1}}
\end{tcolorbox}
\vspace{-1pt}
\begin{tcolorbox}[boxsep=0pt, left=4pt, right=4pt, top=2pt, bottom=2pt,
    colback=blue!4, colframe=blue!35, boxrule=0.3pt, arc=1.5pt]
\textcolor{blue!55!black}{\textbf{Q$_2$}:}\ Who wrote \emph{Animal Farm}?\\
\textcolor{blue!55!black}{\textbf{A$_2$}:}\ George Orwell wrote \emph{Animal Farm}, published in $1945$ -- same author as Q$_1$.\,\textcolor{black!55}{\tiny\textsuperscript{1}}
\end{tcolorbox}
\vspace{-1pt}
\begin{tcolorbox}[boxsep=0pt, left=4pt, right=4pt, top=2pt, bottom=2pt,
    colback=blue!4, colframe=blue!35, boxrule=0.3pt, arc=1.5pt]
\textcolor{blue!55!black}{\textbf{Q$_3$}:}\ Did the author win the Nobel Prize in Literature?\\
\textcolor{blue!55!black}{\textbf{A$_3$}:}\ No, the document lists Orwell's honours (Prometheus Hall of Fame, Retrospective Hugo) and explicitly states he was \emph{not} a Nobel laureate.\,\textcolor{black!55}{\tiny\textsuperscript{2}}
\end{tcolorbox}

\vspace{4pt}\hrule height 0.3pt\vspace{4pt}

\textbf{Verdict}\\[2pt]
\begin{tabular}{@{}r@{\ }l@{}}
\textcolor{green!45!black}{$\boldsymbol{\checkmark}$} &
\textcolor{green!45!black}{\textbf{\method{}}\,: \refuted{}}
\hfill \textcolor{black!55}{\footnotesize Ground Truth: \refuted{}}
\end{tabular}
\end{tcolorbox}
\vspace{-6pt}
\caption{\textbf{Trace for the intro-teaser claim of \cref{fig:intro_teaser}.}
The three sub-questions that survive the useful / informative / diverse filter (Q$_1$--Q$_3$) are each answerable from the evidence, and their composition correctly refutes the claim: both books are Orwell's, but the document's awards section establishes that he was not awarded the Nobel Prize in Literature.
A monolithic classifier or a less-targeted decomposer is likely to be misled by the true premise (``author of two famous novels'') and miss the negated qualifier on the prize -- the failure mode that the necessity reward ($R_\text{nec}$) is designed to surface.}
\label{fig:trace_orwell}
\end{figure*}

\begin{figure*}[t]
\centering
\begin{tcolorbox}[
    width=\textwidth,
    colback=white,
    colframe=black!55,
    boxrule=0.5pt,
    arc=2pt,
    left=6pt, right=6pt, top=5pt, bottom=5pt,
    fontupper=\small,
]
\textbf{Claim} \hfill \textcolor{black!55}{\footnotesize HoVer (2-hop, in-domain)}\\[1pt]
\textit{``Marko Dmitrovi\'c is a player in the La Liga league.''}

\vspace{4pt}\hrule height 0.3pt\vspace{4pt}

\textbf{Evidence}\\[1pt]
\textcolor{black!75}{%
\textsuperscript{1}\,\textbf{Marko Dmitrovi\'c} (born 24 January 1992) is a Serbian professional footballer who plays for Spanish club \emph{SD Eibar} as a goalkeeper.\,
\textsuperscript{2}\,\textbf{Sociedad Deportiva Eibar} is a Spanish football club based in Eibar, Gipuzkoa; the team currently plays in \emph{La Liga}, having been promoted to the top tier of Spanish football in 2014.}

\vspace{4pt}\hrule height 0.3pt\vspace{4pt}

\textbf{\method{} verification trace}\\[2pt]
\vspace{-10pt}
\begin{tcolorbox}[boxsep=0pt, left=4pt, right=4pt, top=2pt, bottom=2pt,
    colback=blue!4, colframe=blue!35, boxrule=0.3pt, arc=1.5pt]
\textcolor{blue!55!black}{\textbf{Q$_1$}:}\ Who plays for SD Eibar?\\
\textcolor{blue!55!black}{\textbf{A$_1$}:}\ Marko Dmitrovi\'c plays for SD Eibar as a goalkeeper.\,\textcolor{black!55}{\tiny\textsuperscript{1}}
\end{tcolorbox}
\vspace{-1pt}
\begin{tcolorbox}[boxsep=0pt, left=4pt, right=4pt, top=2pt, bottom=2pt,
    colback=blue!4, colframe=blue!35, boxrule=0.3pt, arc=1.5pt]
\textcolor{blue!55!black}{\textbf{Q$_2$}:}\ What league does SD Eibar play in?\\
\textcolor{blue!55!black}{\textbf{A$_2$}:}\ SD Eibar currently plays in La Liga.\,\textcolor{black!55}{\tiny\textsuperscript{2}}
\end{tcolorbox}

\vspace{4pt}\hrule height 0.3pt\vspace{4pt}

\textbf{Verdict}\\[2pt]
\begin{tabular}{@{}r@{\ }l@{}}
\textcolor{green!45!black}{$\boldsymbol{\checkmark}$} &
\textcolor{green!45!black}{\textbf{\method{}}\,: \supported{}
{\scriptsize(conf.\ $0.997$)}} \hfill \textcolor{black!55}{\footnotesize Ground Truth: \supported{}}
\end{tabular}
\end{tcolorbox}
\vspace{-6pt}
\caption{\textbf{Multi-hop \supported{} success.}
The model bridges the two atomic facts ``Dmitrovi\'c plays for Eibar'' and ``Eibar plays in La Liga'' through the shared entity \emph{SD Eibar}; the verdict only follows from their conjunction.}
\label{fig:trace_supported}
\end{figure*}

\begin{figure*}[t]
\centering
\begin{tcolorbox}[
    width=\textwidth,
    colback=white,
    colframe=black!55,
    boxrule=0.5pt,
    arc=2pt,
    left=6pt, right=6pt, top=5pt, bottom=5pt,
    fontupper=\small,
]
\textbf{Claim} \hfill \textcolor{black!55}{\footnotesize FEVEROUS (in-domain)}\\[1pt]
\textit{``George Brown began his Liberal Party leadership in Canada on the first of July in 1867.''}

\vspace{4pt}\hrule height 0.3pt\vspace{4pt}

\textbf{Evidence}\\[1pt]
\textcolor{black!75}{%
\textsuperscript{1}\,\textbf{George Brown} reorganized the Clear Grit (Liberal) Party in \emph{1857}, supporting, among other things, the separation of church and state, the annexation of Rupert's Land, and a small government.}

\vspace{4pt}\hrule height 0.3pt\vspace{4pt}

\textbf{\method{} verification trace}\\[2pt]
\vspace{-10pt}
\begin{tcolorbox}[boxsep=0pt, left=4pt, right=4pt, top=2pt, bottom=2pt,
    colback=blue!4, colframe=blue!35, boxrule=0.3pt, arc=1.5pt]
\textcolor{blue!55!black}{\textbf{Q$_1$}:}\ In what year did George Brown reorganize the Liberal Party?\\
\textcolor{blue!55!black}{\textbf{A$_1$}:}\ George Brown reorganized the Liberal Party in 1857.\,\textcolor{black!55}{\tiny\textsuperscript{1}}
\end{tcolorbox}
\vspace{-1pt}
\begin{tcolorbox}[boxsep=0pt, left=4pt, right=4pt, top=2pt, bottom=2pt,
    colback=blue!4, colframe=blue!35, boxrule=0.3pt, arc=1.5pt]
\textcolor{blue!55!black}{\textbf{Q$_2$}:}\ Does the document mention July 1, 1867?\\
\textcolor{blue!55!black}{\textbf{A$_2$}:}\ The document gives 1857 and never mentions July 1, 1867.\,\textcolor{black!55}{\tiny\textsuperscript{1}}
\end{tcolorbox}

\vspace{4pt}\hrule height 0.3pt\vspace{4pt}

\textbf{Verdict}\\[2pt]
\begin{tabular}{@{}r@{\ }l@{}}
\textcolor{green!45!black}{$\boldsymbol{\checkmark}$} &
\textcolor{green!45!black}{\textbf{\method{}}\,: \refuted{}
{\scriptsize(conf.\ $1.000$)}} \hfill \textcolor{black!55}{\footnotesize Ground Truth: \refuted{}}
\end{tabular}
\end{tcolorbox}
\vspace{-6pt}
\caption{\textbf{Clean single-fact \refuted{} success.}
The model isolates the numeric mismatch ($1857$ vs.\ $1867$) with a targeted follow-up question rather than over-relying on the verdict head; the second question explicitly nails the year discrepancy.}
\label{fig:trace_refuted}
\end{figure*}

\begin{figure*}[t]
\centering
\begin{tcolorbox}[
    width=\textwidth,
    colback=white,
    colframe=black!55,
    boxrule=0.5pt,
    arc=2pt,
    left=6pt, right=6pt, top=5pt, bottom=5pt,
    fontupper=\small,
]
\textbf{Claim} \hfill \textcolor{black!55}{\footnotesize FEVEROUS (in-domain)}\\[1pt]
\textit{``Both the Tantalus monkey and the Vervet monkey can be found in Ethiopia and West Africa.''}

\vspace{4pt}\hrule height 0.3pt\vspace{4pt}

\textbf{Evidence}\\[1pt]
\textcolor{black!75}{%
\textsuperscript{1}\,\textbf{Tantalus monkey.} This monkey is found in tropical central Africa.\, Its native range includes Benin, Burkina Faso, Cameroon, the Central African Republic, Chad, Congo, the Democratic Republic of Congo, Ghana, Kenya, Niger, Nigeria, South Sudan, Sudan, Togo, and Uganda.}

\vspace{4pt}\hrule height 0.3pt\vspace{4pt}

\textbf{\method{} verification trace}\\[2pt]
\vspace{-10pt}
\begin{tcolorbox}[boxsep=0pt, left=4pt, right=4pt, top=2pt, bottom=2pt,
    colback=blue!4, colframe=blue!35, boxrule=0.3pt, arc=1.5pt]
\textcolor{blue!55!black}{\textbf{Q$_1$}:}\ What is the native range of the Tantalus monkey?\\
\textcolor{blue!55!black}{\textbf{A$_1$}:}\ Tropical central Africa, including Cameroon, Ghana, Kenya, Nigeria, Sudan, and Uganda.\,\textcolor{black!55}{\tiny\textsuperscript{1}}
\end{tcolorbox}
\vspace{-1pt}
\begin{tcolorbox}[boxsep=0pt, left=4pt, right=4pt, top=2pt, bottom=2pt,
    colback=orange!5, colframe=orange!50, boxrule=0.3pt, arc=1.5pt]
\textcolor{orange!70!black}{\textbf{Q$_2$}:}\ What about the Vervet monkey?\\
\textcolor{orange!70!black}{\textbf{A$_2$}:}\ \emph{I don't know} the range of the Vervet monkey from the given evidence document.
\end{tcolorbox}
\vspace{-1pt}
\begin{tcolorbox}[boxsep=0pt, left=4pt, right=4pt, top=2pt, bottom=2pt,
    colback=blue!4, colframe=blue!35, boxrule=0.3pt, arc=1.5pt]
\textcolor{blue!55!black}{\textbf{Q$_3$}:}\ Is Ethiopia in the Tantalus monkey's listed range?\\
\textcolor{blue!55!black}{\textbf{A$_3$}:}\ No: Ethiopia is not in the listed range, so the claim already fails on the Tantalus side.\,\textcolor{black!55}{\tiny\textsuperscript{1}}
\end{tcolorbox}

\vspace{4pt}\hrule height 0.3pt\vspace{4pt}

\textbf{Verdict}\\[2pt]
\begin{tabular}{@{}r@{\ }l@{}}
\textcolor{green!45!black}{$\boldsymbol{\checkmark}$} &
\textcolor{green!45!black}{\textbf{\method{}}\,: \refuted{}
{\scriptsize(conf.\ $1.000$)}} \hfill \textcolor{black!55}{\footnotesize Ground Truth: \refuted{}}
\end{tabular}
\end{tcolorbox}
\vspace{-6pt}
\caption{\textbf{Calibrated abstention.}
The model abstains on the unsupported Vervet sub-claim (Q$_2$) rather than guessing, then routes the verdict through the answerable Tantalus sub-fact (Q$_3$).
The abstention does not break the verdict because the claim is already refutable from the answerable half, exactly the behaviour the joint-multiplicative reward is designed to elicit (\cref{sec:method:joint}).}
\label{fig:trace_abstention}
\end{figure*}

\begin{figure*}[t]
\centering
\begin{tcolorbox}[
    width=\textwidth,
    colback=white,
    colframe=black!55,
    boxrule=0.5pt,
    arc=2pt,
    left=6pt, right=6pt, top=5pt, bottom=5pt,
    fontupper=\small,
]
\textbf{Claim} \hfill \textcolor{black!55}{\footnotesize CoverBench (out-of-domain, tabular)}\\[1pt]
\textit{``Three out of the five highest-earning golfers on the $2001$ PGA Tour were from the United States.''}

\vspace{4pt}\hrule height 0.3pt\vspace{4pt}

\textbf{Evidence}\\[1pt]
\textcolor{black!75}{%
\textsuperscript{1}\,\textbf{2001 PGA Tour earnings (top 5).}\,
1.\,Tiger Woods (United States);\,
2.\,Phil Mickelson (United States);\,
3.\,David Toms (United States);\,
4.\,Vijay Singh (Fiji);\,
5.\,Davis Love III (United States).}

\vspace{4pt}\hrule height 0.3pt\vspace{4pt}

\textbf{\method{} verification trace}\\[2pt]
\vspace{-10pt}
\begin{tcolorbox}[boxsep=0pt, left=4pt, right=4pt, top=2pt, bottom=2pt,
    colback=blue!4, colframe=blue!35, boxrule=0.3pt, arc=1.5pt]
\textcolor{blue!55!black}{\textbf{Q$_1$}:}\ How many golfers are listed in the table?\\
\textcolor{blue!55!black}{\textbf{A$_1$}:}\ The table lists five golfers.\,\textcolor{black!55}{\tiny\textsuperscript{1}}
\end{tcolorbox}
\vspace{-1pt}
\begin{tcolorbox}[boxsep=0pt, left=4pt, right=4pt, top=2pt, bottom=2pt,
    colback=blue!4, colframe=blue!35, boxrule=0.3pt, arc=1.5pt]
\textcolor{blue!55!black}{\textbf{Q$_2$}:}\ How many of the golfers are from the United States?\\
\textcolor{blue!55!black}{\textbf{A$_2$}:}\ \emph{Three} of the golfers are from the United States: Tiger Woods, Phil Mickelson, and David Toms.\,\textcolor{black!55}{\tiny\textsuperscript{1}}
\hfill \textcolor{red!65!black}{\footnotesize (\textbf{undercount}: omits Davis Love III)}
\end{tcolorbox}
\vspace{-1pt}
\begin{tcolorbox}[boxsep=0pt, left=4pt, right=4pt, top=2pt, bottom=2pt,
    colback=blue!4, colframe=blue!35, boxrule=0.3pt, arc=1.5pt]
\textcolor{blue!55!black}{\textbf{Q$_3$}:}\ What is the order of earnings?\\
\textcolor{blue!55!black}{\textbf{A$_3$}:}\ Woods, Mickelson, Toms, Singh, Love III (descending earnings).\,\textcolor{black!55}{\tiny\textsuperscript{1}}
\end{tcolorbox}

\vspace{4pt}\hrule height 0.3pt\vspace{4pt}

\textbf{Verdict}\\[2pt]
\begin{tabular}{@{}r@{\ }l@{}}
\textcolor{red!65!black}{$\boldsymbol{\times}$} &
\textcolor{red!65!black}{\textbf{\method{}}\,: \supported{}
{\scriptsize(conf.\ $0.998$)}} \hfill \textcolor{black!55}{\footnotesize Ground Truth: \refuted{}}
\end{tabular}
\end{tcolorbox}
\vspace{-6pt}
\caption{\textbf{Counting-style failure on out-of-domain tabular evidence.}
The decomposition is locally sensible (each sub-question is on-topic, and the model retrieves the full ordering in Q$_3$), but Q$_2$ under-counts (the four U.S.\ golfers are 1, 2, 3, \emph{and} 5; Davis Love III at rank $5$ is dropped).
The model then agrees with the claim's stated count of ``three'' instead of cross-checking it against the answer to Q$_3$, yielding a high-confidence \supported{} on a claim whose ground truth label is \refuted{}.
This points to a residual lack of cross-question consistency that none of the seven rewards directly penalizes.}
\label{fig:trace_failure}
\end{figure*}

\FloatBarrier
\newpage
\onecolumn

\section{Prompt Templates}
\label{sec:appendix:prompts}

This section shows the six prompt templates used in \method{}: the user prompt that conditions the policy itself (\cref{sec:appendix:prompt:trace}), the silver-decomposition prompt used during data curation (\cref{sec:appendix:prompt:decompose}), and the four LLM-judge prompts that score reward components during training (\cref{sec:appendix:prompt:question_checker,sec:appendix:prompt:answer_checker,sec:appendix:prompt:atomicity,sec:appendix:prompt:coverage}).

\subsection{Verification Trace Generation}
\label{sec:appendix:prompt:trace}
The user-facing prompt that conditions the \method{} policy to emit a structured verification trace (\texttt{<think>}/\texttt{<question>}/\texttt{<answer>}/\texttt{<verification>} blocks) given a claim and an evidence document.
This is the prompt used for every result reported in the body.

\begin{promptbox}[prompt:trace]{Verification Trace Generation}
\begin{Verbatim}[fontsize=\scriptsize, commandchars=\\\{\}, breaklines=true, breakanywhere=true]
You are tasked with systematically verifying the accuracy of a claim. You will be provided with a claim to verify and an evidence document to consult.

Here is the evidence document you should consult:

<evidence_document>
\promptvar{evidence_doc}
</evidence_document>

Here is the claim you need to verify:

<claim>
\promptvar{claim}
</claim>

Your task is to verify whether this claim is Supported or Refuted through an iterative process of asking questions and gathering information.

\# Verification Process

You will follow an iterative cycle of questioning and reasoning:

\#\# Initial Analysis
Begin by analyzing the claim in <think> tags. In your initial analysis:
- Scan the evidence document for passages that seem potentially relevant to the claim
- Decompose the claim into its atomic sub-claims:
  1. Identify explicit connectives (and, or, but, because, which, etc.) and implicit assumptions, comparisons, or vague terms that each need separate verification
  2. Classify each sub-claim by type (e.g., entity, relational, quantitative, causal, temporal, comparative, etc.)
  3. Note which sub-claims are independently falsifiable -- if any single one is refuted, the entire claim is refuted
- Write out a numbered checklist of these sub-claims (this list will guide your verification cycle)
- Identify any ambiguous, vague, or underspecified elements in the claim
- Determine what specific question you should ask

It's OK for this section to be quite long.

\#\# Iterative Question-Answer Cycle
After your initial analysis, enter an iterative cycle where you:

1. **Ask a Question**: In <question> tags, pose a single specific verification question that addresses one aspect of the claim. Your question should target:
   - A specific atomic sub-claim that needs verification
   - An ambiguous element that needs clarification
   - An underspecified term or concept
   - Any other information needed to determine the claim's accuracy

2. **Answer the Question**: In <answer> tags, answer your question using **only** the evidence document:
   - Search the evidence document for relevant information. If you find relevant passages, quote them directly.
   - If the evidence document contains sufficient information, use it to answer the question and cite the relevant passage.
   - If the evidence document does NOT contain the necessary information, explicitly state "I don't know" and move on. Do NOT use outside knowledge to fill the gap.

3. **Evaluate Sufficiency**: In <think> tags, reason about whether you now have sufficient information to verify the claim. Consider:
   - List which sub-claims have been verified so far and which remain unverified
   - Are there remaining ambiguous or underspecified elements in the claim?
   - Do you need additional information to make a confident verification judgment?
   - If yes to any of these, determine what question to ask next.
   - If no, proceed to final verification.

4. **Repeat or Conclude**:
   - If more information is needed, return to step 1 and ask another question.
   - If you have sufficient information, proceed to final verification.

Continue the cycle until every sub-claim identified in your initial analysis has been addressed. Once all sub-claims are covered, proceed to final verification. Do not ask redundant questions about sub-claims that have already been resolved.

\#\# Final Verification
Once you have gathered sufficient information, provide your final judgment in <verification> tags. Your judgment must be exactly one of these two labels:
- **Supported**: The claim is factually accurate and well-supported by the evidence
- **Refuted**: The claim is factually incorrect or contradicted by the evidence

\# Example Output Structure

Here is an example of the expected output format (with generic placeholder content):

<think>
[Initial analysis of the claim, breaking it down and identifying what needs to be verified first]
</think>

<question>
[First specific verification question]
</question>

<answer>
[Answer based on evidence document only, quoting relevant passages. If not found, state that "I don't know".]
</answer>

<think>
[List which sub-claims are verified, which remain. Determine next question or proceed to final verification.]
</think>

<question>
[Second specific verification question]
</question>

<answer>
[Answer based on evidence document only, quoting relevant passages. If not found, state that "I don't know".]
</answer>

<think>
[List which sub-claims are verified, which remain. Determine next question or proceed to final verification.]
</think>

<question>
[Third specific verification question]
</question>

<answer>
[Answer based on evidence document only, quoting relevant passages. If not found, state that "I don't know".]
</answer>

<think>
[Determination that sufficient information has been gathered to verify the claim]
</think>

<verification>
[Supported OR Refuted]
</verification>

Don't output anything after the final verification tag. Do not include any additional commentary, reasoning, or information beyond the final verification label. Your final output should end immediately after the closing </verification> tag.

Begin your verification process now.
\end{Verbatim}
\end{promptbox}

\subsection{Silver-Decomposition Question Generator}
\label{sec:appendix:prompt:decompose}
Given a claim and its evidence, the judge LLM enumerates the minimal set of atomic sub-questions needed to verify the claim.
The length of the returned list defines $n^\star$ for the silver count target, and the $\geq 2$-question filter in the curation pipeline (\cref{sec:data:filter:decomp}) is applied to this list.

\begin{promptbox}[prompt:decompose]{Silver-Decomposition Question Generator}
\begin{Verbatim}[fontsize=\scriptsize, commandchars=\\\{\}, breaklines=true, breakanywhere=true]
You will be given a claim that needs to be verified and an evidence document to consult. Your task is to generate the minimal set of atomic questions needed to fully verify this claim, where each question can be answered using only the evidence document.

<evidence_document>
\promptvar{evidence_doc}
</evidence_document>

<claim>
\promptvar{claim}
</claim>

Your goal is to produce the minimum set of atomic questions required to verify the claim completely. Consider two types of questions needed:

1. **Atomic sub-claims**: Break down the main claim into its fundamental, indivisible components that each require verification
2. **Under-specified elements**: Identify vague or ambiguous parts of the claim that need clarification to enable proper verification

Guidelines for your analysis:
- Generate between 1 and 20 questions
- Aim for the smallest possible set that still ensures complete verification
- Avoid redundant questions that provide diminishing returns
- Each question must be atomic, so it should address a distinct, necessary aspect of verification
- Consider both factual verification and definitional clarification needs
- Each question must be answerable using only the provided evidence document -- do not generate questions that require external knowledge
- Keep each question concise -- paraphrase rather than quoting long passages from the claim verbatim

Output the list of questions.
\end{Verbatim}
\end{promptbox}

\subsection{Question Answerability Check}
\label{sec:appendix:prompt:question_checker}
Given a document and a candidate question, the judge LLM emits a binary label in $\{0,1\}$ indicating whether the question is fully answerable from the document alone.
Provides $R_\text{ans}^{(i)}$ in the joint multiplicative reward (\cref{sec:method:joint}).

\begin{promptbox}[prompt:question_checker]{Question Answerability Check ($R_\text{ans}$)}
\begin{Verbatim}[fontsize=\scriptsize, commandchars=\\\{\}, breaklines=true, breakanywhere=true]
Determine if a question can be answered using ONLY the provided document.

<document>
\promptvar{document}
</document>

<question>
\promptvar{question}
</question>

\#\# Answerability Criteria

The question is ANSWERABLE (output 1) if:
- The document explicitly states the answer, OR
- The answer can be directly inferred from stated facts

The question is NOT ANSWERABLE (output 0) if:
- The input is not actually a question (e.g., it is a statement, analysis, or explanation)
- The document does not mention relevant information
- The document mentions the topic but lacks specific details needed
- Answering requires external knowledge not in the document

\#\# Important
- If the input is a statement or analysis rather than a question -> Output 0
- "Partially answerable" -> Output 0 (we need FULL answerability)
- If unsure, default to 0

First, briefly explain your reasoning, then provide your final answer inside <answer> tags containing only 0 or 1.
\end{Verbatim}
\end{promptbox}

\subsection{Answer Correctness Check}
\label{sec:appendix:prompt:answer_checker}
Given a document and a candidate answer sentence, the judge LLM emits a binary label in $\{0,1\}$ indicating whether the sentence is consistent with the document and introduces no external facts.
Provides $R_\text{corr}^{(i)}$ in the joint multiplicative reward (\cref{sec:method:joint}).

\begin{promptbox}[prompt:answer_checker]{Answer Correctness Check ($R_\text{corr}$)}
\begin{Verbatim}[fontsize=\scriptsize, commandchars=\\\{\}, breaklines=true, breakanywhere=true]
You are tasked with verifying whether a sentence is correct based solely on the provided document.

<document>
\promptvar{document}
</document>

<sentence>
\promptvar{sentence}
</sentence>

\#\# Verification Rules

Output 1 (CORRECT) if:
- the sentence accurately reflects some information in the document
- The sentence doesn't introduce any information beyond what's in the document
- No factual errors or contradictions with the document

Output 0 (INCORRECT) if:
- The sentence contradicts the document
- The sentence introduces information not found in the document

First, briefly explain your reasoning, then provide your final answer inside <answer> tags containing only 0 or 1.
\end{Verbatim}
\end{promptbox}

\subsection{Question Atomicity Checklist}
\label{sec:appendix:prompt:atomicity}
Given a claim and a candidate question, the judge LLM evaluates the question against five binary atomicity criteria (is-question, single-focus, no-conjunctions, verifiable, grounded).
The average of YES counts defines $R_\text{atom}^{(i)}$ in the joint multiplicative reward (\cref{sec:method:joint}).

\begin{promptbox}[prompt:atomicity]{Question Atomicity Checklist ($R_\text{atom}$)}
\begin{Verbatim}[fontsize=\scriptsize, commandchars=\\\{\}, breaklines=true, breakanywhere=true]
You will evaluate a question against five binary atomicity criteria for verifying a given claim.

<claim>
\promptvar{claim}
</claim>

<question>
\promptvar{question}
</question>

Evaluate the question on each criterion below. Answer YES or NO for each.

1. **Is a question**: Does the text contain an actual question rather than being purely a statement, analysis, or explanation? A brief setup before the question is acceptable, but the text must contain an actual question.
2. **Single-focus**: Does the question ask about exactly one thing? A question fails this if it asks about multiple distinct aspects, facts, or relationships in a single question.
3. **No conjunctions**: Does the question avoid using "and", "or", "as well as", or similar conjunctions to join distinct sub-claims or topics? Minor conjunctions within a single concept (e.g., "cause and effect") are acceptable.
4. **Verifiable**: Does the question have a definitive yes/no or specific factual answer? It should not be open-ended, subjective, or require an essay-length response.
5. **Grounded**: Does the question reference a specific entity, fact, number, or detail from the claim rather than being generic or abstract?

First, briefly reason about each criterion. Then provide your final answers inside <answer> tags in the exact format:

<answer>
is_question:YES/NO
single_focus:YES/NO
no_conjunctions:YES/NO
verifiable:YES/NO
grounded:YES/NO
</answer>
\end{Verbatim}
\end{promptbox}

\subsection{Coverage Verdict from Answers}
\label{sec:appendix:prompt:coverage}
Given a claim and the list of answers produced for its sub-questions, the judge LLM reconstructs the verdict (\supported{} / \refuted{} / \nei{}).
The match against the ground truth label defines the coverage reward $R_\text{cov}$ (\cref{sec:method:rewards}) and is also reused inside the leave-one-out necessity reward $R_\text{nec}$ (\cref{sec:method:necessity}).

\begin{promptbox}[prompt:coverage]{Coverage Verdict from Answers ($R_\text{cov}$)}
\begin{Verbatim}[fontsize=\scriptsize, commandchars=\\\{\}, breaklines=true, breakanywhere=true]
You are tasked with determining the verdict of a claim based on a set of answers to verification questions.

<answers>
\promptvar{answers}
</answers>

<claim>
\promptvar{claim}
</claim>

\#\# Verdict Criteria

**SUPPORTED**: All parts of the claim are confirmed by the answers.
- Every sub-claim has a corresponding information from the answers confirming it
- No contradictions found

**REFUTED**: Any part of the claim is contradicted by an answer.
- At least one answer directly contradicts a sub-claim
- Evidence shows claim is false

**NOT_ENOUGH_INFO**: Answers are insufficient to determine verdict.
- Some sub-claims lack corresponding answers
- Answers are ambiguous or inconclusive

\#\# Process
1. List each sub-claim in the claim
2. Determine if each sub-claim is supported/refuted/unknown based on the answers
3. Aggregate to final verdict

First, briefly explain your reasoning by analyzing how each answer relates to the claim. Then provide your final verdict inside <verdict> tags containing only one of: Supported, Refuted, or Not Enough Information.
\end{Verbatim}
\end{promptbox}

\end{document}